\documentclass{article}

\usepackage{microtype}
\usepackage{graphicx}
\usepackage{subfigure}
\usepackage{booktabs} %

\usepackage{hyperref}
\usepackage{soul}
\usepackage[accepted]{icml2020}

\usepackage{amsmath,amsthm,amsfonts,amssymb}
\usepackage{enumerate,color,xcolor}
\usepackage{graphicx}
\usepackage{subfigure}
\usepackage{url}

\usepackage{enumitem}

\usepackage{multirow}

\theoremstyle{plain}

\newtheorem{theorem}{Theorem}

\newtheorem{corollary}{Corollary}

\newtheorem{proposition}{Proposition}

\newtheorem{assumption}{Assumption}

\usepackage{comment}

\newcommand{\xb}{\mathbf{x}}
\newcommand{\vb}{\mathbf{v}}
\newcommand{\zb}{\mathbf{z}}
\newcommand{\rset}{\mathbb{R}}

\newcommand{\rmd}{\mathrm{d}}
\newcommand{\Bm}{\mathrm{B}}
\newcommand{\Lm}{\mathrm{L}^\alpha}

\newcommand{\beq}{\begin{eqnarray}}
\newcommand{\eeq}{\end{eqnarray}}
\newcommand{\beqs}{\begin{eqnarray*}}
\newcommand{\eeqs}{\end{eqnarray*}}
\newcommand{\sas}{{\cal S}\alpha{\cal S}}

\DeclareMathOperator*{\argmin}{arg\min}

\usepackage{mathtools}

\usepackage{silence}
\ActivateWarningFilters[pdftoc]
\ActivateWarningFilters[xclr]

\icmltitlerunning{Fractional Underdamped Langevin Dynamics}
\begin{document}

\twocolumn[
\icmltitle{Fractional Underdamped Langevin Dynamics: \\Retargeting SGD with Momentum under Heavy-Tailed Gradient Noise}

\icmlsetsymbol{equal}{*}

\begin{icmlauthorlist}
\icmlauthor{Umut \c{S}im\c{s}ekli}{equal,tpt,ox}
\icmlauthor{Lingjiong Zhu}{equal,fsu}
\icmlauthor{Yee Whye Teh}{ox}
\icmlauthor{Mert G\"{u}rb\"{u}zbalaban}{ru}
\end{icmlauthorlist}

\icmlaffiliation{tpt}{LTCI, T\'{e}l\'{e}com Paris, Institut Polytechnique de Paris, Paris, France}
\icmlaffiliation{ox}{Department of Statistics, University of Oxford, Oxford, UK}
\icmlaffiliation{fsu}{Department of Mathematics, Florida State University, Tallahassee, USA}
\icmlaffiliation{ru}{Department of Management Science and Information Systems, Rutgers Business School, Piscataway, USA}

\icmlcorrespondingauthor{Umut \c{S}im\c{s}ekli}{umut.simsekli@telecom-paris.fr}

\icmlkeywords{Stochastic Gradient Descent, Stochastic Differential Equations, Kinetic Langevin, Deep Learning}

\vskip 0.3in
]

\printAffiliationsAndNotice{\icmlEqualContribution} %

\begin{abstract}
Stochastic gradient descent with momentum (SGDm) is one of the most popular optimization algorithms in deep learning. While there is a rich theory of SGDm for convex problems, the theory is considerably less developed in the context of deep learning where the  problem is non-convex and the gradient noise might exhibit a heavy-tailed behavior, as empirically observed in recent studies. In this study, we consider a \emph{continuous-time} variant of SGDm, known as the underdamped Langevin dynamics (ULD), and investigate its asymptotic properties under heavy-tailed perturbations. Supported by recent studies from statistical physics, we argue both theoretically and empirically that the heavy-tails of such perturbations can result in a bias even when the step-size is small, in the sense that \emph{the optima of stationary distribution} of the dynamics might not match \emph{the optima of the cost function to be optimized}. As a remedy, we develop a novel framework, which we coin as \emph{fractional} ULD (FULD), and prove that FULD targets the so-called Gibbs distribution, whose optima exactly match the optima of the original cost. We observe that the Euler discretization of FULD has noteworthy algorithmic similarities with \emph{natural gradient} methods and \emph{gradient clipping}, bringing a new perspective on understanding their role in deep learning. We support our theory with experiments conducted on a synthetic model and neural networks.

\end{abstract}

\section{Introduction}

Gradient-based optimization algorithms have been the de facto choice in deep learning for solving the optimization problems of the form:
\begin{align}
\xb^\star = \argmin\nolimits_{\xb \in \rset^d} \Bigl\{ f(\xb) \triangleq (1/n) \sum\nolimits_{i=1}^n f^{(i)}(\xb) \Bigr\}\,, \label{eqn:optim}
\end{align}
where $f:\rset^d\to\rset$ denotes the non-convex loss function, $f^{(i)}$ denotes the loss contributed by an individual data point $i \in \{1, \dots, n\}$, $\xb\in\rset^d$ denotes the collection of all the parameters of the neural network.
Among others, stochastic gradient descent with momentum (SGDm) is one of the most popular algorithms for solving such optimization tasks (see e.g., \citet{sutskever2013importance,iclr2018}), and is based on the following iterative scheme:
\begin{align}
\tilde{\vb}^{k+1}= \tilde{\gamma} \tilde{\vb}^k - \tilde{\eta} \nabla \tilde{f}_{k+1}(\xb^{k}), \quad \xb^{k+1} = \xb^{k} + \tilde{\vb}^{k+1}, \label{eqn:sgdm_common}
\end{align} 
where $k$ denotes the iteration number, $\tilde{\eta}$ is the step-size, $\tilde{\gamma}$ is the friction, and $\tilde{\vb}$ denotes the \emph{velocity} (also referred to as momentum). Here, $\nabla\tilde{f}_k$ denotes the stochastic gradients defined as follows:
\begin{align}
\nabla\tilde{f}_k(\xb) \triangleq (1/b) \sum\nolimits_{i\in \Omega_k} f^{(i)}(\xb), \label{eqn:stochgrad}
\end{align}
where $\Omega_k \subset \{1,\dots,n\}$ denotes a random subset drawn from the set of data points with $|\Omega_k| = b \ll n$ for all $k$. 

When the gradients are computed on all the data points (i.e., $\nabla \tilde{f}_k = \nabla f$), SGDm becomes \emph{deterministic} and can be viewed as a discretization of the following \emph{continuous-time} system \cite{GGZ,maddison2018hamiltonian}:
\begin{align}
\rmd \vb_t=- (\gamma \vb_t+\nabla f(\xb_t)) \rmd t, \qquad \rmd \xb_t= \vb_t \rmd t, \label{eqn:ode}
\end{align}
where $\vb_t$ is still called the velocity. The connection between this system and \eqref{eqn:sgdm_common} becomes clearer, if we discretize this system by using the Euler scheme with step-size $\eta$:
\begin{align}
\nonumber &\vb^{k+1}= \vb^{k} - \eta (\gamma \vb^{k+1} +\nabla f(\xb^{k}))\,,
\\
&\xb^{k+1} = \xb^{k} + \eta \vb^{k+1}\,, \label{eqn:sgdm}
\end{align}
and make the change of variables $\tilde{\vb}^k \triangleq \eta \vb^k$, $\tilde{\gamma} \triangleq (1-\eta \gamma) $, and $\tilde{\eta} \triangleq \eta^2$. %
However, due to the presence of the stochastic gradient noise $U_k(\xb) \triangleq \nabla\tilde{f}_k(\xb) - \nabla f(\xb)$, the sequence $\{\xb_k, \vb_k\}_{k\in \mathbb{N}_+}$ will be a \emph{stochastic process} and the deterministic system \eqref{eqn:ode} would not be an appropriate proxy.

Understanding the statistical properties of $\{\xb_k, \vb_k\}_{k\in \mathbb{N}_+}$ would be of crucial importance as it might reveal the peculiar properties that lie behind the performance of SGDm for learning with neural networks. 
A popular approach for understanding the dynamics of stochastic optimization algorithms in deep learning is to impose some structure on the noise $U_k$ and relate the process \eqref{eqn:sgdm_common} to a stochastic differential equation (SDE) \cite{mandt2016variational,jastrzkebski2017three,hu2017diffusion,chaudhari2018stochastic,zhu2019anisotropic,pmlr-v97-simsekli19a}. For instance, by assuming that the second-order moments of the stochastic gradient noise are bounded (i.e., $\mathbb{E}\|U_k(\xb)\|^2 < \infty$ for all admissible $k$, $\xb$), one might argue that $U_k$ can be approximated by a Gaussian random vector due to the central limit theorem (CLT) \cite{fischer2011history}. Under this assumption, we might view \eqref{eqn:sgdm_common} as a discretization of the following SDE, which is also known as the \emph{underdamped} or \emph{kinetic} Langevin dynamics:
\begin{align}
\nonumber &\rmd \vb_t=- (\gamma \vb_t+\nabla f(\xb_t))\rmd t + \sqrt{2\gamma/\beta} \rmd \Bm_t
\\
&\rmd \xb_t= \vb_t \rmd t,\label{eqn:sde_brownian}
\end{align} 
where $\Bm_t$ denotes the $d$-dimensional Brownian motion and $\beta > 0$ is called the inverse temperature variable, measuring the noise intensity along with $\gamma$. It is easy to check that, under very mild assumptions, the solution process $\{\xb_t,\vb_t\}_{t\geq 0}$ admits an invariant distribution whose density is proportional to $\exp(-\beta(f(\xb) + \|\vb\|^2/2))$, where the function $\|\vb\|^2/2$ is often called the \emph{Gaussian kinetic energy} (see e.g. \cite{betancourt2017geometric}) and the distribution itself is called the Boltzmann-Gibbs measure \cite{pavliotis2014stochastic,GGZ,herau2004isotropic,dalalyan2018kinetic}. We then observe that the marginal distribution $\xb$ in the stationarity has a density proportional to $\exp(-\beta f(\xb))$, which indicated that any local minimum of $f$ appears as a local maximum of this density. This is a desirable property since it implies that, when the gradient noise $U_k$ has light tails, the process will spend more time near the local minima of $f$. Furthermore, it has been shown that as $\beta$ goes to infinity, the marginal distribution of $\xb$ concentrates around the global optimum $\xb^\star$. This observation has yielded interesting results for understanding the dynamics of SGDm in the contexts of both sampling and optimization with convex and non-convex potentials $f$ \cite{GGZ,GGZ2,pmlr-v80-zou18a,lu2016relativistic,csimcsekli2018asynchronous}.

\begin{figure}[t]
    \centering
    \includegraphics[width=0.49\columnwidth]{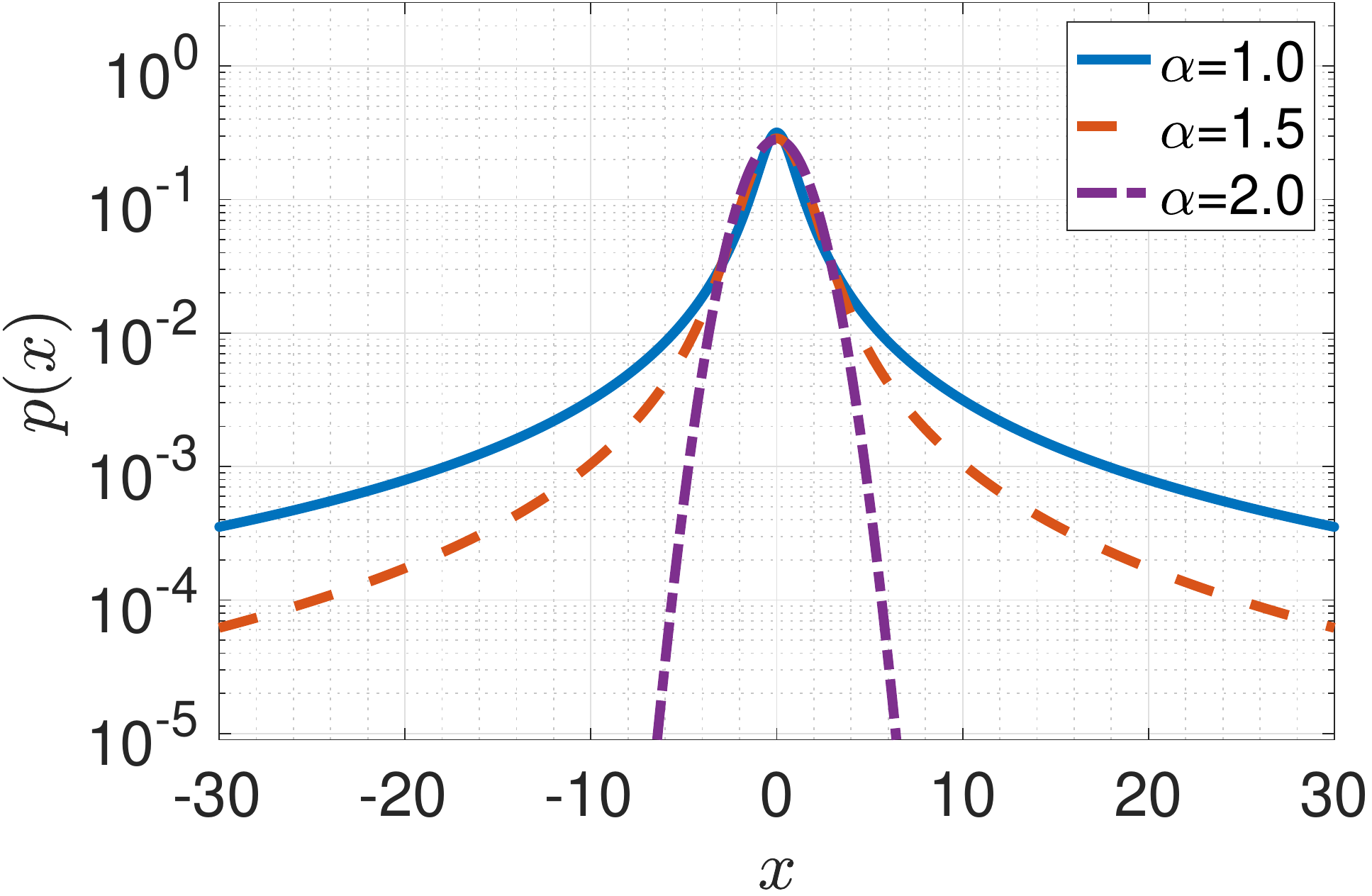} \hfill
    \includegraphics[width=0.49\columnwidth]{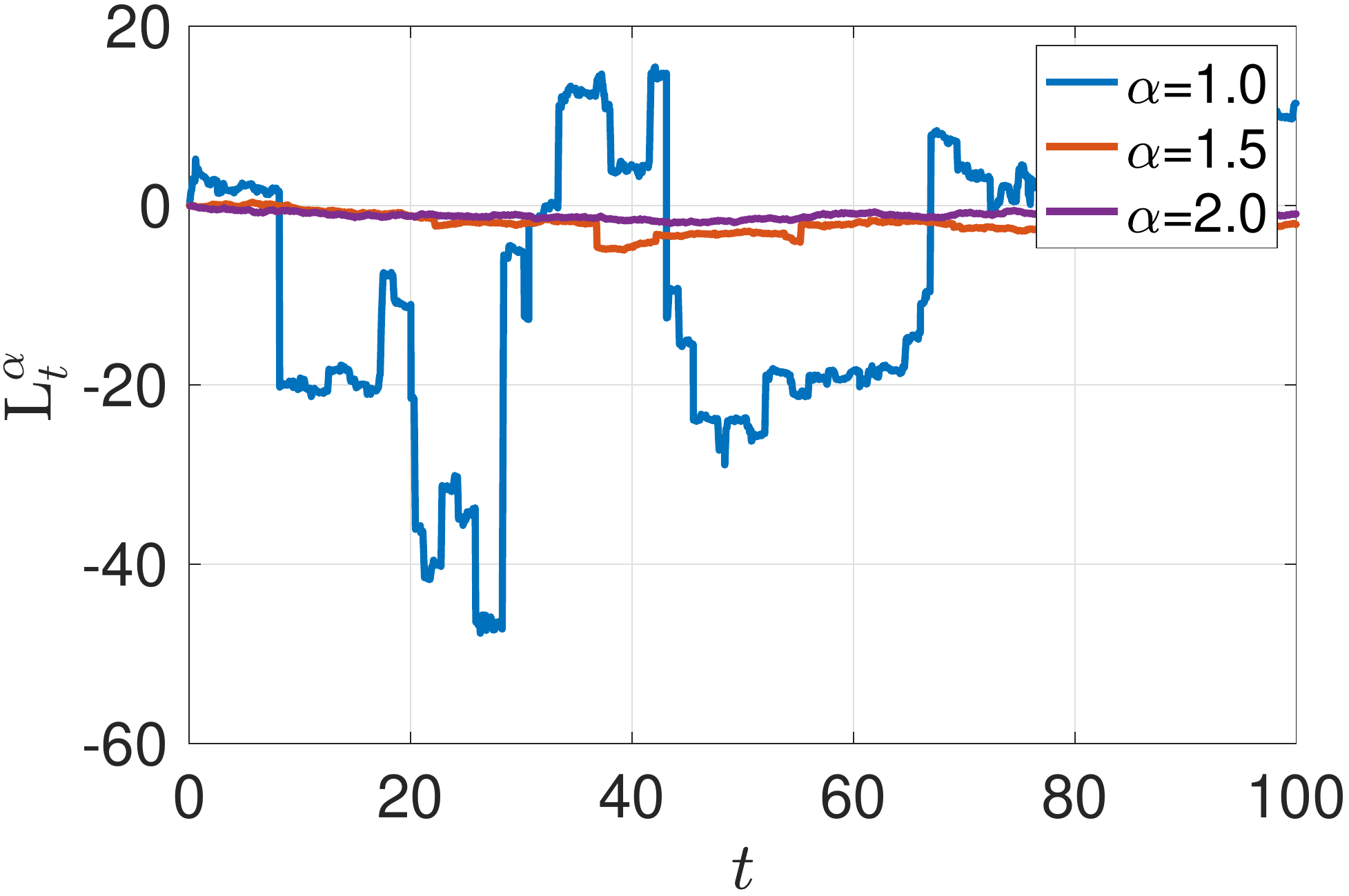} 
    \caption{$\sas$ densities and $\Lm_t$.}
    \label{fig:sas}
\end{figure}

While the Gaussianity assumption can be accurate in certain settings such as small networks \cite{martin2019heavy,panigrahi2019non}, recently it has been empirically demonstrated that in several deep learning setups, the stochastic gradient noise can exhibit a \emph{heavy-tailed} behavior \cite{csimcsekli2019heavy,zhang2019adam}\footnote{In two recent studies, \citet{gurbuzbalaban2020heavy} and \citet{hodgkinson2020multiplicative} have shown that
the stationary distribution of the stochastic gradient descent (SGD) algorithm can be indeed a heavy-tailed distribution depending on the choice of the step-size and the batch-size. On the other hand, in another recent study, \citet{csimcsekli2020hausdorff} have provided generalization bounds for a general class of SDEs, including heavy- and light-tailed ones.}. While the Gaussianity assumption would not be appropriate in this case since the conventional CLT would not hold anymore, nevertheless we can invoke the generalized CLT, which states that the asymptotic distribution of $U_k$ will be a symmetric $\alpha$-stable distribution 
($\mathcal{S}\alpha\mathcal{S}$); a class of distributions that are commonly used in the statistical physics literature as an approximation to heavy-tailed random variables \citep{sliusarenko2013stationary,dubkov2008levy}. As we will define in more detail in the next section, in the core of $\sas$, lies the parameter $\alpha \in (0,2]$, which determines the heaviness of the tail of the distribution. The tails get heavier as $\alpha$ gets smaller, the case $\alpha=2$ reduces to the Gaussian random variables. This is illustrated in Figure \ref{fig:sas}.

{\citet{pmlr-v97-simsekli19a,csimcsekli2019heavy} empirically illustrated that, in deep neural networks, the statistical structure of $U_k$ can be better captured by using an $\alpha$-stable distribution.}  With the assumption of $U_k$ being $\sas$ distributed, the choice of Brownian motion will be no longer appropriate and should be replaced with an $\alpha$-stable L\'{e}vy motion, which motivates the following L\'{e}vy-driven SDE:
\begin{align}
\nonumber  &\rmd \vb_t=- (\gamma \vb_{t-}+\nabla f(\xb_t))\rmd t + \sqrt{2\gamma/\beta} \rmd \Lm_t,
\\
&\rmd \xb_t= \vb_t \rmd t,\label{eqn:sde_sas}
\end{align} 
where $\vb_{t-}$ denotes the left limit of $\vb_t$ and $\Lm_t$ denotes the $\alpha$-stable L\'{e}vy process with independent components, which coincides with $\sqrt{2}\Bm_t$ when $\alpha=2$. Unfortunately, when $\alpha<2$, as opposed to its Brownian counterpart, the invariant measures of such SDEs do not admit an analytical form in general; yet, one can still show that the invariant measure cannot be in the form of the Boltzmann-Gibbs measure \cite{eliazar2003levy}.

A more striking property of \eqref{eqn:sde_sas} was very recently revealed in a statistical physics study \cite{capala2019stationary}, where the authors numerically illustrated that, even when $f$ has a single minimum, the invariant measure of \eqref{eqn:sde_sas} can exhibit multiple maxima, none of which coincides with the minimum of $f$. A similar property has been formally proven in the overdamped dynamics with Cauchy noise (i.e., $\alpha=1$ and $\gamma \to \infty$) by \citet{sliusarenko2013stationary}. 
Since the process \eqref{eqn:sde_sas} would spend more time around the modes of its invariant measure (i.e., the high probability region), in an optimization context (i.e., for larger $\beta$) the sample paths would concentrate around these modes, which might be arbitrarily distant from the optima of $f$. In other words, the heavy-tails of the gradient noise could result in an undesirable bias, which would be still present even when the step-size is taken to be arbitrarily small.
As we will detail in Section~\ref{sec:mainres}, informally, this phenomenon stems from the fact that the heavy-tailed noise leads to aggressive updates on $\vb$, which are then directly transmitted to $\xb$ due to the dynamics. Unless `tamed', these updates create an hurling effect on $\xb$ and drift it away from the modes of the ``potential" $f$ that is sought to be minimized.

\textbf{Contributions:} In this study, we develop a \emph{fractional} underdamped Langevin dynamics whose invariant distribution is guaranteed to be in the form of the Boltzmann-Gibbs measure, hence its optima exactly match the optima of $f$. 
We first prove a general theorem which holds for any kinetic energy function, which is not necessarily the Gaussian kinetic energy. However, it turns out that some components of the dynamics might not admit an analytical form for an arbitrary choice of the kinetic energy. Then we identify two choices of kinetic energies, where all the terms in dynamics can be written in an analytical form or accurately computable. We also analyze the Euler discretization of \eqref{eqn:sde_levy} and identify sufficient conditions for ensuring weak convergence of the ergodic averages computed over the iterates.

We observe that the discretization of the proposed dynamics has interesting algorithmic similarities with natural gradient descent \cite{amari1998natural} and gradient clipping \citep{pascanu2013difficulty}, which we believe bring further theoretical understanding for their role in deep learning. Finally, we support our theory with experiments conducted on both synthetic settings and neural networks.

\section{Technical Background \& Related Work}

The stable distributions are heavy-tailed
distributions
that appear as the limiting distribution of the generalized CLT for
a sum of i.i.d. random variables
with infinite variance \cite{paul1937theorie}.
In this paper, we are interested in 
centered \textit{symmetric $\alpha$-stable 
distribution}.
A scalar random variable $X$
follows a symmetric $\alpha$-stable 
distribution denoted as $X\sim\mathcal{S}\alpha\mathcal{S}(\sigma)$ 
if its characteristic function
takes the form:
$\mathbb{E}\left[e^{i\omega X}\right]=\exp\left(-\sigma^{\alpha}|\omega|^{\alpha}\right)$, $\omega\in\mathbb{R}$,
where $\alpha\in(0,2]$ and $\sigma>0$.
Here, $\alpha\in(0,2]$ is known as the tail-index, which determines
the tail thickness of the distribution.
$\mathcal{S}\alpha\mathcal{S}$ becomes
heavier-tailed as $\alpha$ gets smaller.
$\sigma>0$ is known as the scale
parameter that measures the spread
of $X$ around $0$.
The probability density function of a symmetric $\alpha$-stable distribution, $\alpha\in(0,2]$,
does not yield closed-form expression in general except for a few special cases.
When $\alpha=1$ and $\alpha=2$, $\sas$ reduces to the Cauchy and the Gaussian distributions, respectively.
When $0<\alpha<2$, $\alpha$-stable distributions 
have heavy-tails so that their moments
are finite only up to the order $\alpha$
in the sense that 
$\mathbb{E}[|X|^{p}]<\infty$
if and only if $p<\alpha$,
which implies infinite variance.

L\'{e}vy motions are stochastic processes with independent and stationary increments.
Their successive displacements are random 
and independent, and statistically identical over different time intervals of the same length,
and can be viewed as the continuous-time
analogue of random walks.
The best known and most important examples
are the Poisson process, Brownian motion,
the Cauchy process and more generally stable
processes. L\'{e}vy motions
are prototypes of Markov processes and
of semimartingales, and concern many 
aspects of probability theory.
We refer to \cite{bertoin1996}
for a survey on the theory of L\'{e}vy motions.

In general, L\'{e}vy motions 
are heavy-tailed, which make it appropriate
to model natural phenomena with possibly
large variations, {that often occurs}
in statistical physics \cite{eliazar2003levy},
signal processing \cite{kuruoglu1999},
and finance \cite{mandelbrot2013}.

We define $\Lm_{t}$, a $d$-dimensional 
symmetric $\alpha$-stable L\'{e}vy motion
with independent components as follows. 
Each component of $\Lm_{t}$ is an independent scalar $\alpha$-stable L\'{e}vy process,
which is defined as follows: (cf.\ Figure~\ref{fig:sas})
\begin{enumerate}[label=(\roman*),noitemsep,topsep=0pt,leftmargin=*,align=left]
\item
$\Lm_{0}=0$ almost surely.
\item
For any $t_{0}<t_{1}<\cdots<t_{N}$, the increments $\Lm_{t_{n}}-\Lm_{t_{n-1}}$
are independent, $n=1,2,\ldots,N$.
\item
The difference $\Lm_{t}-\Lm_{s}$ and $\Lm_{t-s}$
have the same distribution: $\mathcal{S}\alpha\mathcal{S}((t-s)^{1/\alpha})$ for $s<t$.
\item
$\Lm_{t}$ has stochastically continuous sample paths, i.e.
for any $\delta>0$ and $s\geq 0$, $\mathbb{P}(|\Lm_{t}-\Lm_{s}|>\delta)\rightarrow 0$
as $t\rightarrow s$.
\end{enumerate}
When $\alpha=2$,
we obtain a scaled Brownian motion $\sqrt{2}\Bm_{t}$
as a special case so that
the difference $\Lm_{t}-\Lm_{s}$
follows a Gaussian distribution $\mathcal{N}(0,2(t-s))$ and $\Lm_{t}$
is almost surely continuous.
When $0<\alpha<2$, due to the stochastic
continuity property, symmetric $\alpha$-stable
L\'{e}vy motions can have
have a countable number of discontinuities, 
which are often known as \textit{jumps}.
The sample paths are continuous from 
the right and they have left limits,
a property known as c\`{a}dl\`{a}g \cite{duan2015}.

Recently, \citet{FLMC} extended the \emph{overdamped} Langevin dynamics to an SDE driven by $\Lm_t$, given as:\footnote{In \citet{FLMC}, \eqref{eqn:flmc} does not contain an inverse temperature $\beta$, which was later on introduced in \citet{nguyen19}. }
\begin{equation}
\rmd \xb_{t}=b(\xb_{t-},\alpha)\rmd t+ \beta^{-1/\alpha} \rmd\Lm_{t}, \label{eqn:flmc}
\end{equation}
where the drift $b(\xb,\alpha)=((b(\xb,\alpha))_{i},1\leq i\leq d)$ is defined as follows:
\begin{equation}
(b(\xb,\alpha))_{i}={-\mathcal{D}_{x_{i}}^{\alpha-2}(\phi(\xb)\partial_{x_{i}}f(\xb))}/{\phi(\xb)}.
\end{equation}
Here, $\phi(\xb)=\exp(-f(\xb))$ and $\mathcal{D}$ denotes the fractional Riesz derivative \cite{Riesz}:
\begin{equation}
\mathcal{D}^{\gamma}u(x):=\mathcal{F}^{-1}\left\{|\omega|^{\gamma}(\mathcal{F}(u))(\omega)\right\}(x),
\end{equation}
where $\mathcal{F}$ denotes the Fourier transform. Briefly, $\mathcal{D}^\gamma$ extends usual differentiation to fractional orders and when $\gamma =2$ it coincides (up to a sign difference) with the usual second-order derivative $- \rmd^2 f(x)/ \rmd x^2$. 

The important property of the process \eqref{eqn:flmc} is that it admits an invariant distribution whose density is proportional to $\exp(-\beta f(\xb))$ \cite{nguyen19}. It is easy to show that, when $\alpha=2$, the drift reduces to $b(\xb,2) = -\nabla f(\xb)$, hence we recover the classical overdamped dynamics:
\begin{align}
\rmd \xb_{t}=-\nabla f(\xb_t)\rmd t+ \sqrt{2/\beta} \rmd\Bm_{t}. \label{eqn:langevin}
\end{align}

Since the fractional Riesz derivative is costly to compute, 
\citet{FLMC} proposed an approximation of $b(\xb,\alpha)$ based on the alternative definition of $\mathcal{D}$ given in \cite{ortigueira2006riesz}, such that:
\begin{equation}
b(\xb,\alpha)\approx-c_{\alpha}\nabla f(\xb), \label{eqn:riesz_approx}
\end{equation}
where $c_{\alpha}:=\Gamma(\alpha-1)/\Gamma(\alpha/2)^{2}$. This approximation essentially results in replacing $\Bm_t$ with $\Lm_t$ in \eqref{eqn:langevin} in a rather straightforward manner.  
While avoiding the computational issues originated from the Riesz derivatives, as shown in \cite{nguyen19}, this approximation can induce an arbitrary bias in a non-convex optimization context. Besides, the stationary distribution of this approximated dynamics was analytically derived in \cite{sliusarenko2013stationary} under the choice of $\alpha=1$ and $f(x) = x^4/4 - a x^2/2$ for $x\in\rset^1$ and $a>0$. These results show that, in the presence of heavy-tailed perturbations, the drift should be modified, otherwise an inaccurate approximation of the Riesz derivatives can result in an explicit bias, which moves the modes of the distribution away from the modes of $f$.

From a pure Monte Carlo perspective, \citet{SFHMC} extended the fractional overdamped dynamics \eqref{eqn:flmc} to higher-order dynamics and proposed the so-called fractional Hamiltonian dynamics (FHD), given as follows:
\begin{align}
\nonumber \rmd \xb_{t}=&{\mathcal{D}^{\alpha-2}\{\phi(\zb_{t})\vb_{t}\}}/{\phi(\zb_{t})}\rmd t,
\\
\nonumber \rmd \vb_{t}=&-{\mathcal{D}^{\alpha-2}\{\phi(\zb_{t})\nabla f(\xb_{t})\}}/{\phi(\zb_{t})}\rmd t \\
&-\gamma{\mathcal{D}^{\alpha-2}\{\phi(\zb_{t})\vb_{t}\}}/{\phi(\zb_{t})}\rmd t
+\gamma^{1/\alpha}\rmd \Lm_{t}, \label{eqn:fhd}
\end{align}
where $\zb_{t}=(\xb_{t},\vb_{t})$, and $\phi(\zb)=e^{-f(\xb)-\frac{1}{2}\Vert\vb\Vert^{2}}$. They showed that the invariant measure of the process has a density proportional to 
$\phi(\zb)$, i.e., the Boltzmann-Gibbs measure. Similar to the overdamped case \eqref{eqn:flmc}, the Riesz derivatives do not admit an analytical form in general. Hence they approximated them by using the same approximation given in \eqref{eqn:riesz_approx}, which yields the SDE given in \eqref{eqn:sde_sas} (up to a scaling factor). This observation also confirms that the heavy-tailed noise requires an adjustment in the dynamics, otherwise the induced bias might drive the dynamics away from the minima of $f$ \cite{capala2019stationary}.  

\section{Fractional Underdamped Langevin Dynamics}
\label{sec:mainres}

In this section, we develop the fractional underdamped Langevin dynamics (FULD), which is expressed by the following SDE:
\begin{align}
\nonumber &\rmd \vb_t=- (\gamma c(\vb_{t-},\alpha)+\nabla f(\xb_t)) \rmd t+ ({\gamma}/{\beta})^{1/\alpha} \rmd \Lm_t,
\\
&\rmd \xb_t =\nabla g(\vb_t) \rmd t, \label{eqn:sde_levy}
\end{align}
where $c : \rset^d \times (0,2] \mapsto \rset^d$ is the \emph{drift function} for the velocity and $g : \rset^d \mapsto \rset$ denotes a general notion of \emph{kinetic energy}. In the next theorem, which is the main theoretical result of this paper, we will identify the relation between these two functions such that the solution process will keep the \emph{generalized} Boltzmann-Gibbs measure, $\exp(-\beta (f(\xb) + g(\vb)))\rmd\xb\rmd\vb$ invariant. All the proofs are given in the supplementary document.
\begin{theorem}\label{thm:invariant}
Let $c(\vb,\alpha)=((c(\vb,\alpha))_{i}, 1\leq i\leq d)$ has the following form:
\begin{equation}
(c(\vb,\alpha))_{i}:=\frac{\mathcal{D}_{v_{i}}^{\alpha-2}(\psi(\vb)\partial_{v_{i}}g(\vb))}{\psi(\vb)},
\qquad
\psi(\vb):=e^{-g(\vb)}. 
\label{eqn:driftc}
\end{equation}
{The measure} $\pi(\rmd\xb,\rmd\vb)\propto e^{- \beta(f(\xb)+g(\vb))}\rmd\xb \rmd\vb$ on $\mathbb{R}^{d}\times\mathbb{R}^{d}$ is an invariant probability measure for the Markov process $(\xb_{t},\vb_{t})$.
\end{theorem}
One of the main features of FULD is that the fractional Riesz derivatives only appears in the drift $c$, which \emph{only} depends on $\vb$. This is highly in contrast with FHD \eqref{eqn:fhd}, where the Riesz derivatives are taken over both $\xb$ and $\vb$, which is the source of intractability.  
Moreover, FULD enjoys the freedom to choose different kinetic energy functions $g(\vb)$. In the sequel, we will investigate two options for $g$, such that the drift $c$ can be analytically obtained.

\subsection{Gaussian kinetic energy}

In classical overdamped Langevin dynamics and Hamiltonian dynamics, the default choice of kinetic energy is the Gaussian kinetic energy, which corresponds to taking $g(\vb)=\frac{1}{2}\Vert \vb\Vert^{2}$ \cite{neal2010mcmc,livingstone2019kinetic,dalalyan2018kinetic}. With this choice, the fractional dynamics becomes:
\begin{align}
&\rmd\vb_{t}=-\gamma c(\vb_{t-},\alpha)\rmd t-\nabla f(\xb_{t})\rmd t+(\gamma/\beta)^{1/\alpha}\rmd \Lm_{t},
\nonumber
\\
&\rmd \xb_{t}=\vb_{t}\rmd t. \label{eqn:fuld_v1}
\end{align}
In the next result, we will show that in this case, the drift $c$ admits an analytical solution.
\begin{theorem}\label{prop:formula}
Let $g(\vb)=\frac{1}{2}\Vert \vb\Vert^{2}$. Then, for any $1\leq i\leq d$,
\begin{equation}
(c(\vb,\alpha))_{i}
=\frac{2^{\frac{\alpha}{2}}v_{i}}{\sqrt{\pi}}\Gamma\left(\frac{\alpha+1}{2}\right)
{_1F}_{1}\left(\frac{2-\alpha}{2};\frac{3}{2};\frac{v_{i}^{2}}{2}\right),
\end{equation}
where $\Gamma$ is the gamma function and $_{1}F_{1}$ is
the Kummer confluent hypergeometric function.
In particular, when $\alpha=2$, we have $(c(\vb,\alpha))_{i}=v_{i}$.
\end{theorem}
We observe that the fractional dynamics \eqref{eqn:fuld_v1} strictly extends the underdamped Langevin dynamics \eqref{eqn:sde_brownian} as $c(\vb,2)= \vb$.

\begin{figure}[t]
    \centering
    \subfigure[]{\includegraphics[width=0.495\columnwidth]{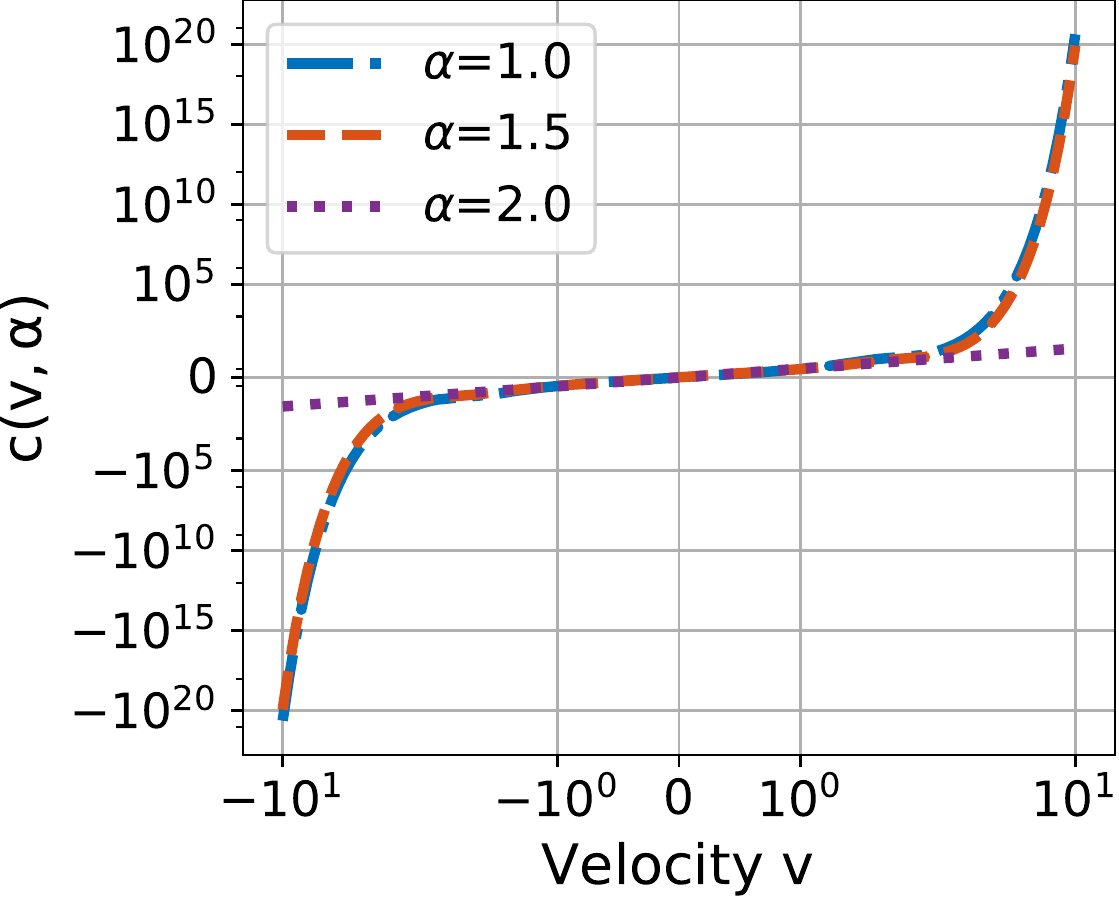} \label{fig:driftv1}}
    \subfigure[]{\includegraphics[width=0.46\columnwidth]{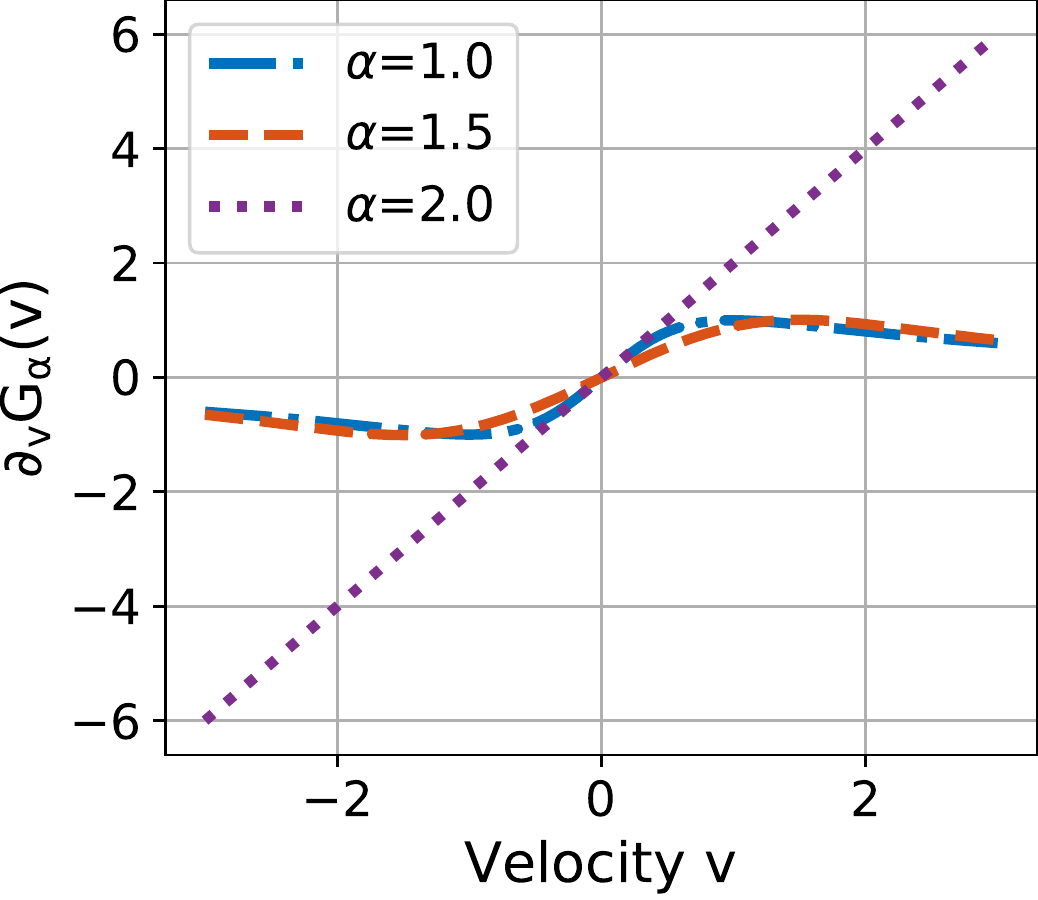} \label{fig:driftv2}}
    \vspace{-10pt}
    \caption{Illustration of one dimensional  a) drift function $c$ for the Gaussian kinetic energy, b) $\nabla G_\alpha$ for the $\sas$ kinetic energy. }

\end{figure}

Let us now investigate the form of the new drift $c$ and its implications. In Figure~\ref{fig:driftv1}, we illustrate $c$ for the $d=1$ dimensional case (note that for $d>1$, each component of $c$ still behaves like Figure~\ref{fig:driftv1}). We observe that due to the hypergeometric function $_{1}F_{1}$, the drift grows exponentially fast with $|v|$ whenever $\alpha<2$. Semantically, this means that, in order to be able to compensate the large jumps incurred by $\Lm_t$, the drift has to react very strongly and hence prevent $v$ to take large values. To illustrate this behavior, we provide more visual illustrations in the supplementary document.

Even though this aggressive behavior of $c$ can be beneficial for the continuous-time system, it is unfortunately clear that its Euler-Maruyama discretization will not yield a practical algorithm due to the same behavior. Indeed, we would need the function $c$ to be Lipschitz continuous in order to guarantee the algorithmic stability of its discretization \cite{kloeden2013numerical}; however, if we consider the integral form of $_{1}F_{1}$ (cf.\ \cite{AS1972}), we observe that the function
\begin{equation*}
(c(\vb,\alpha))_{i}
=\frac{2^{\frac{\alpha}{2}}v_{i}}{\sqrt{\pi}}
\cdot\frac{\Gamma(\frac{3}{2})}{\Gamma(\frac{2-\alpha}{2})}
\int_{0}^{1}e^{\frac{v_{i}^{2}}{2}t}t^{-\frac{\alpha}{2}}(1-t)^{\frac{\alpha-1}{2}}\rmd t
\end{equation*}
is clearly not Lipschitz continuous in $v_{i}$. Therefore, we conclude that FULD with the Gaussian kinetic energy is mostly of theoretical interest.

\subsection{Alpha-stable kinetic energy}

The dynamics with the Gaussian kinetic energy requires a very strong drift $c$ 
mainly because we force the dynamics to make sure that the invariant distribution of $\vb$ to be a Gaussian. Since the Gaussian distribution has light-tails, it cannot tolerate samples with large magnitudes, hence requires a large dissipation to make sure $\vb$ does not take large values.

In order to avoid such an explosive drift that potentially degrades practicality, next we explore \emph{heavy-tailed} kinetic energies, which would allow the components of $\vb$ to take large values, while still making sure that the drift $c$ in \eqref{eqn:driftc} admits an analytical form.   

In our next result, we show that, when we choose an $\sas$ kinetic energy, such that the tail-index $\alpha$ of this kinetic energy matches the one of the driving process $\Lm_t$, the drift $c$ simplifies and becomes the identity function.
\begin{theorem}\label{prop:v}
Let $e^{-g_{\alpha}(v)}$ be the probability density
function of $\mathcal{S}\alpha\mathcal{S}(\frac{1}{\alpha^{1/\alpha}})$.
Choose $\psi(\vb)=e^{-G_{\alpha}(\vb)}$ in \eqref{eqn:driftc}, 
where $G_{\alpha}(\vb)=\sum_{i=1}^{d}g_{\alpha}(v_{i})$ for any $\vb=(v_{1},\ldots,v_{d})$.
Then, 
\begin{equation}
(c(\vb,\alpha))_{i}=v_{i},\qquad 1\leq i\leq d.
\end{equation}
\end{theorem}
This result hints that, perhaps $G_\alpha(v)$ is the natural choice of kinetic energy for the systems driven by $\Lm_t$.

It now follows from Theorem \ref{prop:v} that
the FULD with $\alpha$-stable kinetic energy reduces to the following SDE:
\begin{align}
&\rmd \vb_{t}=-\gamma \vb_{t-}\rmd t-\nabla f(\xb_{t})\rmd t+(\gamma/\beta)^{1/\alpha}\rmd\Lm_{t},
\nonumber
\\
&\rmd \xb_{t}=\nabla G_{\alpha}(\vb_{t})\rmd t.
\label{eqn-sde}
\end{align}
It can be easily verified that $\nabla G_{\alpha}(\vb_{t})=\vb_{t}$ for $\alpha=2$, as $g_{2}(v)=\frac1{2}\log2\pi+\frac1{2}v^2$, hence, the SDE \eqref{eqn-sde} also reduces to the classical underdamped Langevin dynamics \eqref{eqn:sde_brownian}.

While this choice of $g$ results in an analytically available $c$, unfortunately the function $\nabla G_{\alpha}$ itself admits a closed-form analytical formula only when $\alpha=1$ or $\alpha=2$, due to the properties of the $\sas$ densities. Nevertheless, as $\nabla G_{\alpha}$ is based on one-dimensional $\sas$ densities, it can be very accurately computed by using the recent methods developed in \cite{Ament2017}. {On the other hand, in the next section, we will show that $\nabla G_{\alpha}$ is Lipschitz continuous for all $\alpha \in (0,2]$, which implies that under standard regularity conditions on $f$, the Boltzmann-Gibbs measure is the unique invariant measure of \eqref{eqn-sde}. }

We visually inspect the behavior of $\nabla G_\alpha$ in Figure~\ref{fig:driftv2} for dimension one. We observe that, as soon as $\alpha <2$, $\nabla G_\alpha$ takes a very smooth form. Besides, for small $|v|$ the function behaves like a linear function and when $|v|$ goes to infinity, it vanishes. This behavior can be interpreted as follows: since $\vb$ can take larger values due to the heavy tails of the kinetic energy, in order to be able target the correct distribution, the dynamics compensates the potential bursts in $\vb$ by passing it through the asymptotically vanishing $\nabla G_\alpha$. 

\subsection{Euler discretization and weak convergence analysis}

As visually hinted in Figure~\ref{fig:driftv2}, the function $\nabla G_\alpha$ has strong regularity, which makes \eqref{eqn-sde} to be potentially beneficial for practical implementations. Indeed, it is easy to verify that $\nabla G_\alpha$ is Lipschitz continuous for $\alpha =1$ and $2$, and in our next result, we show that this observation is true for any admissible $\alpha$, which is a desired property when discretizing continuous-time dynamics.
\begin{proposition}
\label{prop:lipschitz}
For $0<\alpha\leq 2$, the map $v\mapsto g'_\alpha(v)$ is Lipschitz continuous, hence $\vb\mapsto \nabla G_\alpha(\vb)$ is also Lipschitz continuous. 
\end{proposition}
Accordingly we consider the following Euler-Maruyama discretization for \eqref{eqn-sde}:
\begin{align}
\textstyle
\vb^{k+1} &= \tilde{\gamma}_{k} \vb^k - \eta_{k}  \nabla f(\xb^k) + (\eta_{k} \gamma/\beta)^{1/\alpha}  \mathbf{s}^{k+1}, \nonumber \\
\xb^{k+1} &= \xb^{k} + \eta_{k}\nabla G_\alpha(\vb^{k+1}), \label{eqn:eulerv2}
\end{align}
where $\tilde{\gamma}_{k} = 1-  \gamma \eta_{k}$, $\mathbf{s}^{k}$ is a random vector whose components are independently $\sas(1)$ distributed, and  $(\eta_k)_{k \in \mathbb{N}_+ }$ is a sequence of step-sizes.

In this section, we analyze the weak convergence of the ergodic averages computed by using \eqref{eqn:eulerv2}. Given a test function $h$, consider its expectation with respect to the target measure $\pi$, i.e. $\pi(h) := \mathbb{E}_{X \sim \pi} [h(X)]=\int h(\xb) \pi(\rmd\xb)$ with $\pi(\rmd \xb) \propto \exp(-\beta f(\xb))\rmd \xb$. We will discuss next how this expectation can be approximated through the sample averages  
\begin{equation} 
\bar{\pi}_K(h):= (1/S_K)\sum\nolimits_{k=1}^K \eta_k h(\xb^{k})\,, \label{eqn:mc_avr}
\end{equation} 
where $S_K := \sum_{k=1}^K \eta_k $ is the cumulative sum of the step-size sequence. 

{ We note that Langevin-based algorithms have been used in the literature
to obtain global convergence guarantees for non-convex optimization, see e.g. \cite{Raginsky,xu2018global,GGZ2,ZXG2019,nguyen19}. 
In particular, \citet{nguyen19} used an overdamped
fractional Langevin dynamics for non-convex optimizations.
The proposed model in our paper can also be used to 
study the non-convex optimizations 
and we expect that our underdamped dynamics may
have improved theoretical guarantees compared to \cite{nguyen19}.}

We now present the assumptions that imply our results.
\begin{assumption}\label{assump-stepsize} The step-size sequence $\{\eta_k\}$ is non-increasing and satisfies $\lim_{k\to\infty}\eta_k = 0$ and $\lim_{K\to\infty} S_K =\infty$. 
\end{assumption}
\begin{assumption}\label{assump-growth} Let $V:\mathbb{R}^{2d}\to\mathbb{R}_+$ be a twice continuously differentiable function, satisfying $\lim_{\|\zb\|\to\infty} V(\zb)=\infty$, $\|\nabla V\|\leq C \sqrt{V}$ for some $C>0$ and has a bounded Hessian $\nabla^2 V$. Given $p\in (0,\frac{1}{2}]$, there exists $a\in (1-\frac{p}{2},1]$, $\beta_1 \in\mathbb{R}$, $\beta_2 > 0$ such that $\|b\|^2 \leq CV^a$ and $\langle \nabla V, b \rangle \leq \beta_1 - \beta_2 V^a$ where $b(\vb,x)= (-\gamma \vb - \nabla f(\xb), \nabla G_\alpha(\vb))$ is the drift of the $(\vb_{t},\xb_{t})$ process defined in \eqref{eqn-sde}. 
\end{assumption}
These are common assumptions ensuring that the SDE is simulated with infinite time-horizon and the process is not explosive \cite{panloup2008recursive,FLMC}. We can now establish the weak convergence of \eqref{eqn:mc_avr} and present it as a corollary to Theorem~\ref{thm:invariant}, Proposition~\ref{prop:lipschitz}, and \cite{panloup2008recursive} (Theorem 2).
\begin{corollary} 
\label{cor:weakconv}
Assume that the gradient $\nabla f$ is Lipschitz continuous and has linear growth i.e., there exists $C>0$ such that $\|\nabla f(\xb)\| \leq C(1+ \|\xb\|)$ for all $\xb$. Furthermore, assume that Assumptions \ref{assump-stepsize} and \ref{assump-growth} hold for some $p\in(0,1/2]$. If the test function $h = o(V^{\frac{p}{2}+a-1})$ then
$$\bar{\pi}_K(h) \to \pi(h) \quad \mbox{almost surely as }K\to\infty. $$
\end{corollary}

\subsection{Connections to existing approaches}

\label{sec:conn}

We now point out interesting algorithmic connections between \eqref{eqn:eulerv2} and two methods that are commonly used in practice. We first roll back our initial hypothesis that the gradient noise is $\sas$ distributed, i.e., $\nabla \tilde{f}_{k}(\xb) = \nabla f(\xb) + (\eta_{k} \gamma/\beta)^{1/\alpha}  \mathbf{s}^{k} $, and modify \eqref{eqn:eulerv2} as follows:
\begin{align}
\textstyle
\vb^{k+1} &= \tilde{\gamma}_{k} \vb^k - \eta_{k}  \nabla \tilde{f}_{k+1}(\xb^k), \nonumber \\
\xb^{k+1} &= \xb^{k} + \eta_{k}\nabla G_\alpha(\vb^{k+1}). \label{eqn:eulerv2_optim}
\end{align}
{As a special case when $\tilde{\gamma}_k=0$, we obtain a stochastic gradient descent-type recursion: 
\begin{align}
\xb^{k+1} = \xb^{k} + \eta_{k} \nabla G_\alpha(-\eta_{k}  \nabla \tilde{f}_{k+1}(\xb^k)). \label{eqn:newsgd}
\end{align}}
Let us now consider \emph{gradient-clipping}, a heuristic approach for eliminating the problem of `exploding gradients', which often appear in training neural networks \cite{pascanu2013difficulty,zhang2019analysis}. Very recently, \citet{zhang2019adam} empirically illustrated that such explosions stem from heavy-tailed gradients and formally proved that gradient clipping indeed improves convergence rates under heavy-tailed perturbations. We notice that, the behavior of \eqref{eqn:eulerv2_optim} is reminiscent of gradient clipping: due to the vanishing behavior of $\nabla G_\alpha$ for $\alpha<2$, as the components of $\vb^k$ gets larger in magnitude, the update applied on $\xb^k$ gets smaller. The behavior becomes more prominent in \eqref{eqn:newsgd}. On the other hand, \eqref{eqn:eulerv2_optim} is more aggressive in the sense that the updates can get arbitrarily small as the value of $\alpha$ decreases as opposed to being `clipped' with a threshold.

The second connection is with the natural gradient descent algorithm, where the stochastic gradients are pre-conditioned with the inverse Fisher information matrix (FIM) \cite{amari1998natural}. Here FIM is defined as $\mathbb{E}[\nabla f(\xb)\nabla f(\xb)^\top]$, where the expectation is taken over the data. Notice that when $\alpha=1$ (i.e., Cauchy distribution), we have the following form: $\nabla G_{1}(\vb)=\left(\frac{2v_{1}}{v_{1}^{2}+1},\ldots,\frac{2v_{d}}{v_{d}^{2}+1}\right)$. Therefore, we observe that, in \eqref{eqn:newsgd}, $\nabla G_{1}(\nabla \tilde{f}_k(\xb))$ can be equivalently written as $\mathbf{M}_k(\xb)^{-1}\nabla \tilde{f}_k(\xb)$, where $\mathbf{M}_k(\xb)$ is a diagonal matrix with entries $m_{ii} = ((\nabla \tilde{f}_k(\xb))_i^2 +1)/2$.
Therefore, we can see $\mathbf{M}_k$ as an estimator of the diagonal part of FIM, as they will be in the same order when $|(\nabla \tilde{f}_k(\xb))_i|$ is large.
Besides, \eqref{eqn:eulerv2_optim} then appears as its momentum extension. 
However, $\mathbf{M}_k$ will be biased mainly due to the fact that FIM is the average of the squared gradients, whereas $\mathbf{M}_k$ is based on the square of the average gradients. This connection is rather surprising, since a seemingly unrelated, differential geometric approach turns out to have strong algorithmic similarities with a method that naturally arises when the gradient noise is Cauchy distributed.

 \begin{figure}[t]

    \centering
    \includegraphics[width=0.99\columnwidth]{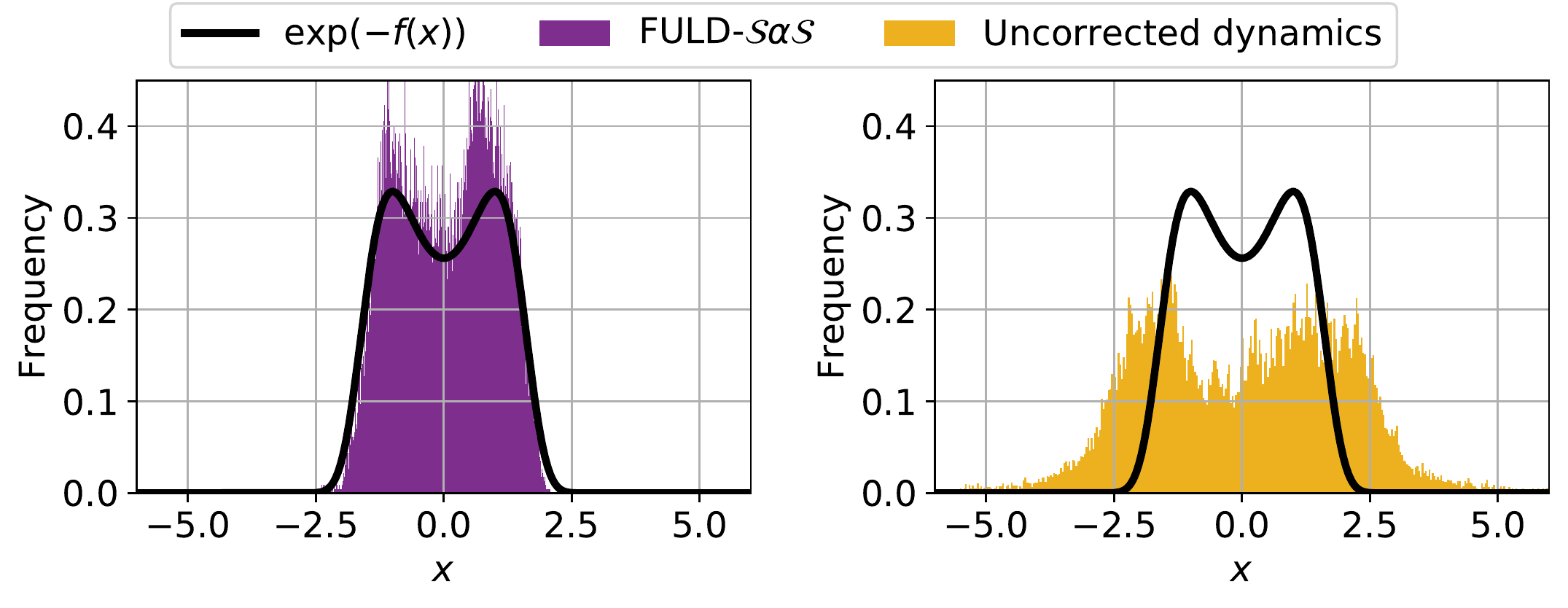}\\
    \includegraphics[width=0.99\columnwidth]{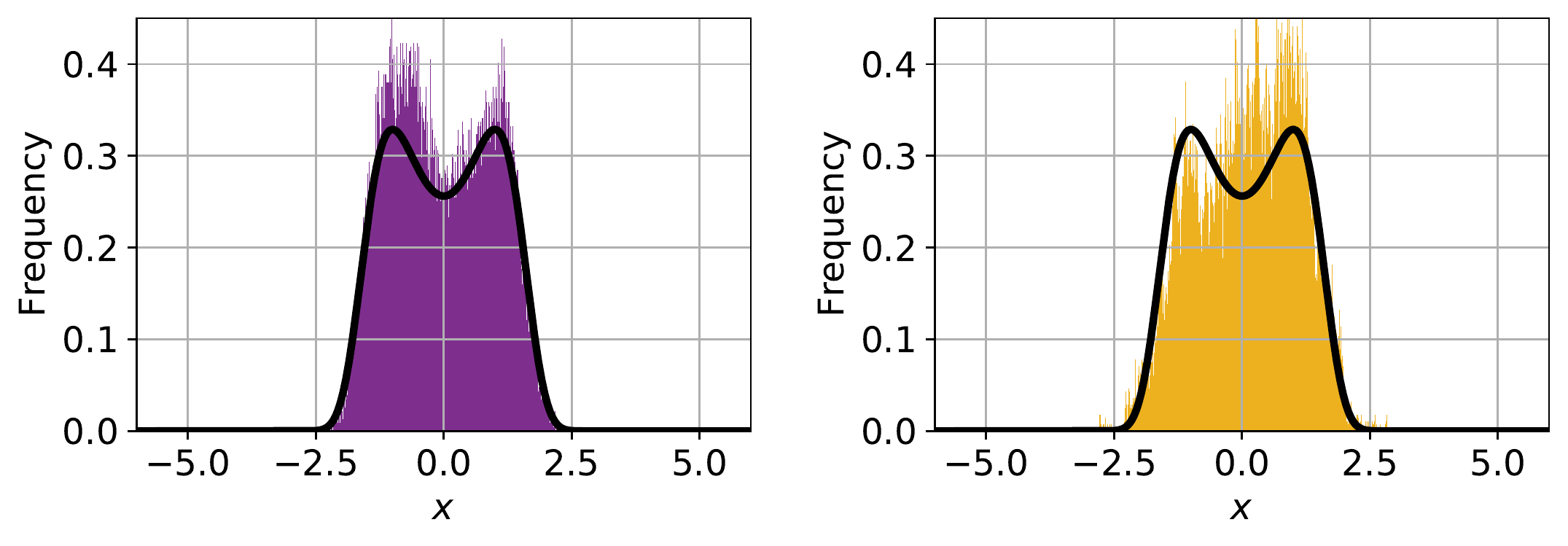}
    \vspace{-15pt}
    \caption{Estimated invariant measures for the quartic potential: top $\alpha=1$, bottom $\alpha = 1.9$.}
    \label{fig:synth}
\end{figure}

\section{Numerical Study}

In this section, we will illustrate our theory on several experiments which are conducted in both synthetic and real-data settings\footnote{We provide our implementation in \url{https://github.com/umutsimsekli/fuld}.}. We note that, as expected, FULD with Gaussian kinetic energy did not yield a numerically stable discretization due to the explosive behavior of $c$. Hence, in this section, we only focus on FULD with $\sas$ kinetic energy and from now on we will simply refer to FULD with $\sas$ kinetic energy as FULD. %

\subsection{Synthetic setting}

 \begin{figure}[t]

    \centering
    \includegraphics[width=0.99\columnwidth]{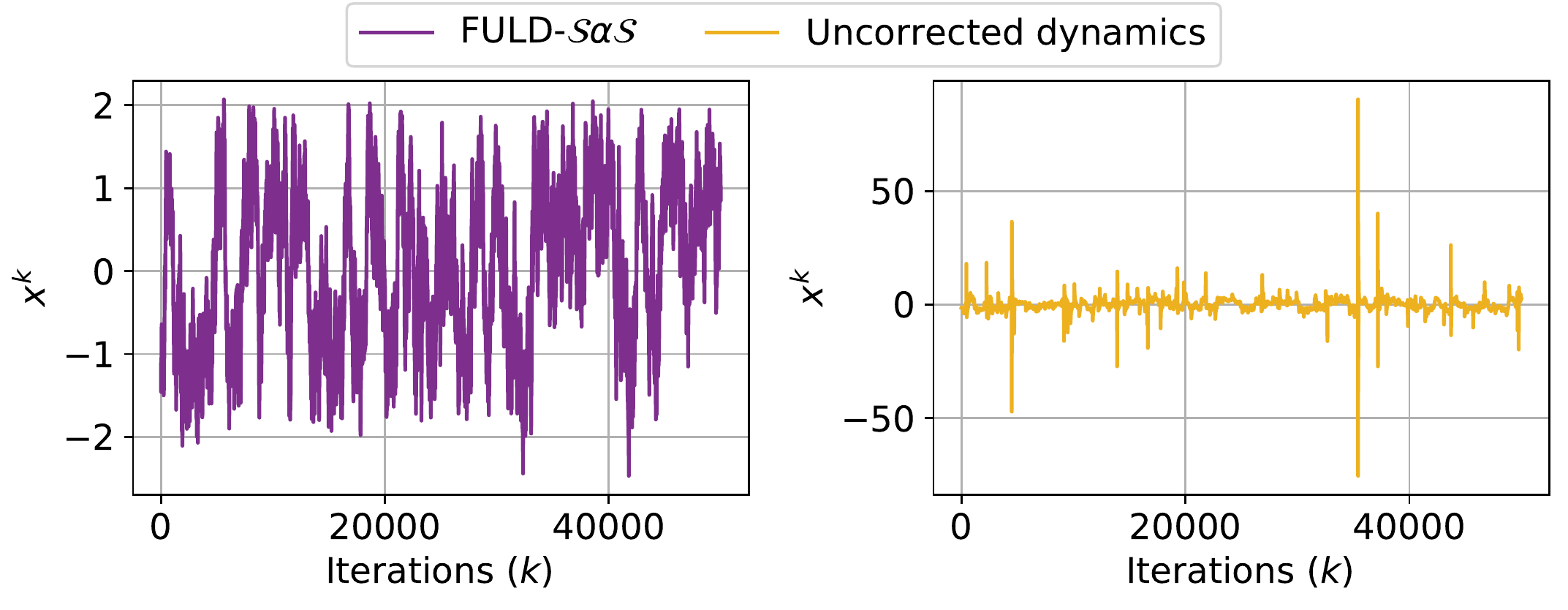}\\
    \includegraphics[width=0.99\columnwidth]{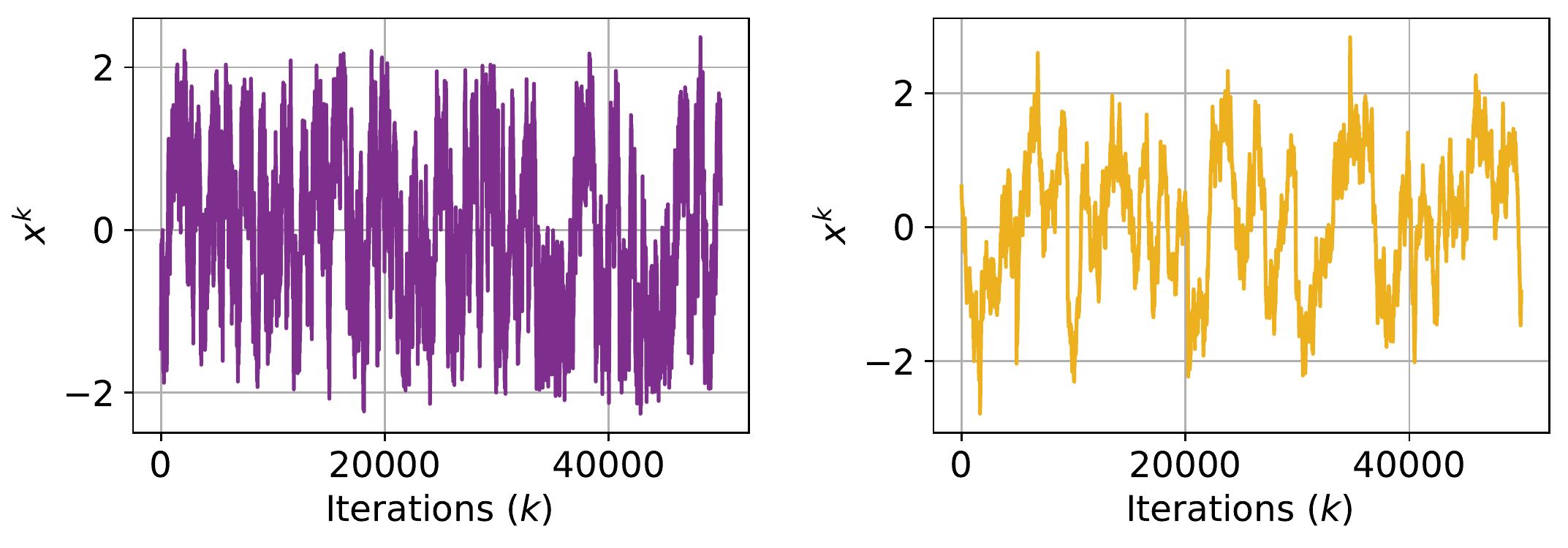}
    \vspace{-13pt}
    \caption{Illustration of the iterates for the quartic potential: top $\alpha=1$, bottom $\alpha = 1.9$.}
    \label{fig:synth_path}
\end{figure}

We first consider a one-dimensional synthetic setting, similar to the one considered in \cite{capala2019stationary}. We consider a quartic potential function with a quadratic component, $f(x) = x^4/4 - x^2/2$. We then simulate the `uncorrected dynamics' (UD) given in \eqref{eqn:sde_sas} and FULD \eqref{eqn-sde} by using the Euler-Maruyama discretization to compare their behavior for different $\alpha$. For $\alpha \notin \{1,2\}$, we used the software given in \cite{Ament2017} for computing $\nabla G_\alpha$.   

Figure~\ref{fig:synth} illustrates the distribution of the samples generated by simulating the two dynamics. In this setup, we set $\beta =1$, $\eta = 0.01$, $\gamma=10$ with number of iterations $K=50000$. We observe that, for $\alpha=1.9$, FULD very accurately captures the form of the distribution, whereas UD exhibits a visible bias and the shape of its resulting distribution is slightly distorted. Nevertheless, since the perturbations are close to a Gaussian in this case (i.e., $\alpha$ is close to $2$), the difference is not substantial and can be tolerable in an optimization context. However, this behavior becomes much more emphasized when we use a heavier-tailed driving process: when $\alpha=1$, we observe that the target distribution of UD becomes distant from the Gibbs measure $\exp(-f(x))$, and more importantly its modes no longer match the minima of $f$; agreeing with the observations presented in \cite{capala2019stationary}\footnote{{We note that the overdamped dynamics with the uncorrected drift exhibits a similar behavior to the one of the uncorrected underdamped dynamics with sufficiently large $\gamma$.} }. On the other hand, thanks to the correction brought by $\nabla G_\alpha$, FULD still captures the target distribution very accurately, even when the driving force is Cauchy.   

On the other hand, in our experiments we observed that, for small values of $\alpha$, UD can quickly become numerically unstable and even diverge for slightly larger step-sizes, whereas this problem never occurred for FULD. This outcome also stems from the fact that UD does not have any mechanism to compensate the potential large updates originating from the heavy-tailed perturbations. To illustrate this observation more clearly, in Figure~\ref{fig:synth_path} we illustrate the iterates $(\xb^k)_{k=1}^K$ which were used for producing Figure~\ref{fig:synth}. We observe that, while the iterates of UD are well-behaved for $\alpha=1.9$, the magnitude range of the iterates gets quite large when $\alpha$ is set to $1$. On the other hand, for both values of $\alpha$, FULD iterates are always kept in a reasonable range, thanks to the clipping-like effect of $\nabla G_\alpha$.

\subsection{Neural networks}

In our next set of experiments, we evaluate our theory on neural networks. In particular, we apply the iterative scheme given in \eqref{eqn:eulerv2_optim} as an optimization algorithm for training neural networks, and compare its behavior with classical SGDm defined in \eqref{eqn:sgdm_common}. In this setting, we do not add any explicit noise, all the stochasticity comes from the potentially heavy-tailed stochastic gradient noise \eqref{eqn:stochgrad} {under the assumption that the noise can be well-modeled by using an $\sas$ vector (see Section~\ref{sec:conn} for the explicit assumption)}.

  \begin{figure}[t]

    \centering
    \includegraphics[width=0.99\columnwidth]{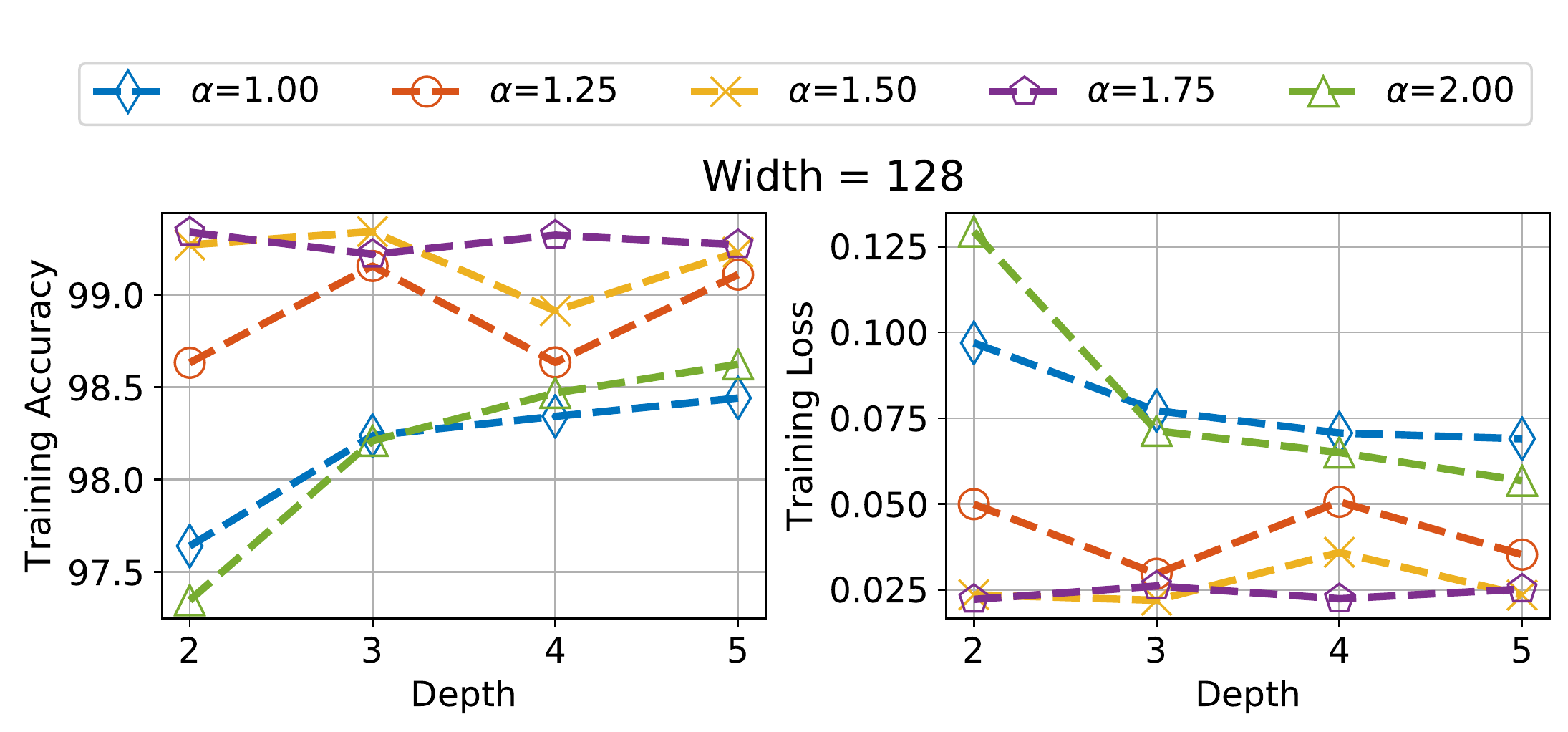}\\
    \includegraphics[width=0.99\columnwidth]{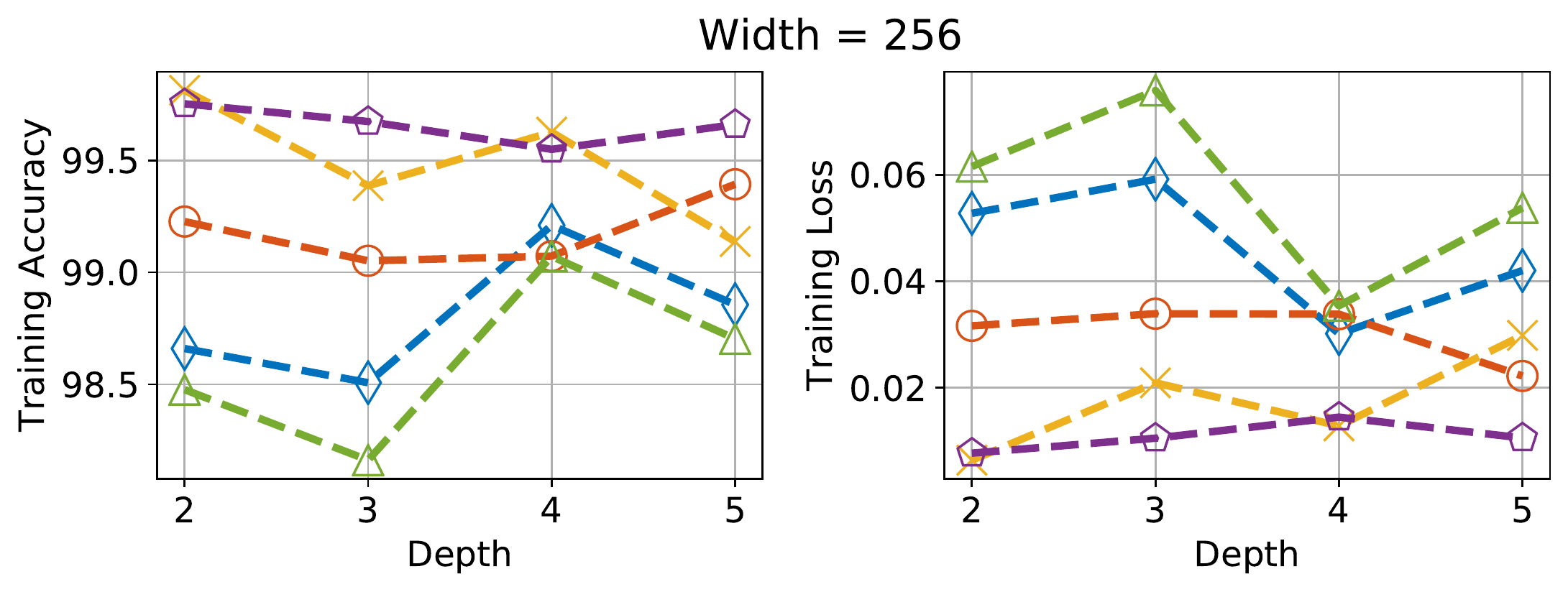}
    \vspace{-10pt}
    \caption{Neural network results on MNIST (training).}
    \label{fig:fcn_mnist}
\end{figure}

We consider a fully-connected network for a classification task on the MNIST and CIFAR10 datasets, with different depths (i.e.\ number of layers) and widths (i.e.\ number of neurons per layer).
For each depth-width pair, we train two neural networks by using SGDm \eqref{eqn:sgdm_common} and our modified version \eqref{eqn:eulerv2_optim}, and compare their final train/test accuracies and loss values. We use the conventional train-test split of the datasets: for MNIST we have $60$K training and $10$K test samples, and for CIFAR10 these numbers are $50$K and $10$K, respectively. We use the cross entropy loss (also referred to as the `negative-log-likelihood'). 

We note that the modified scheme \eqref{eqn:eulerv2_optim} reduces to \eqref{eqn:sgdm_common} when $\alpha =2$, since $\nabla G_2(\vb) = \vb$. Hence in this section, we will refer to SGDm as the special case of \eqref{eqn:eulerv2_optim} with $\alpha =2$. 
On the other hand, in these experiments, directly computing $\nabla G_\alpha$ becomes impractical for $\alpha \notin \{1,2\}$, since the algorithms given in \cite{Ament2017} become prohibitively slow with the increased dimension $d$. However, since $\nabla G_\alpha$ is based on the derivatives of the \emph{one-dimensional} $\sas$ densities $g_\alpha(v)$ (see Theorem~\ref{prop:v}), for $\alpha \in (1,2)$, we first precomputed the values of $g_\alpha(v)$ over a fine grid of $v \in [-100,100]$; then, during the SGDm recursion, we approximated $\nabla G_\alpha$ by linearly interpolating the values of $g_\alpha$ that are precomputed over this grid. 
We expect that, if the stochastic gradient noise can be well-approximated by using an $\sas$ distribution, then the modified dynamics should exhibit an improved performance since it would eliminate the potential bias brought by the heavy-tailed noise.  

 \begin{figure}[t]

    \centering
    \includegraphics[width=0.99\columnwidth]{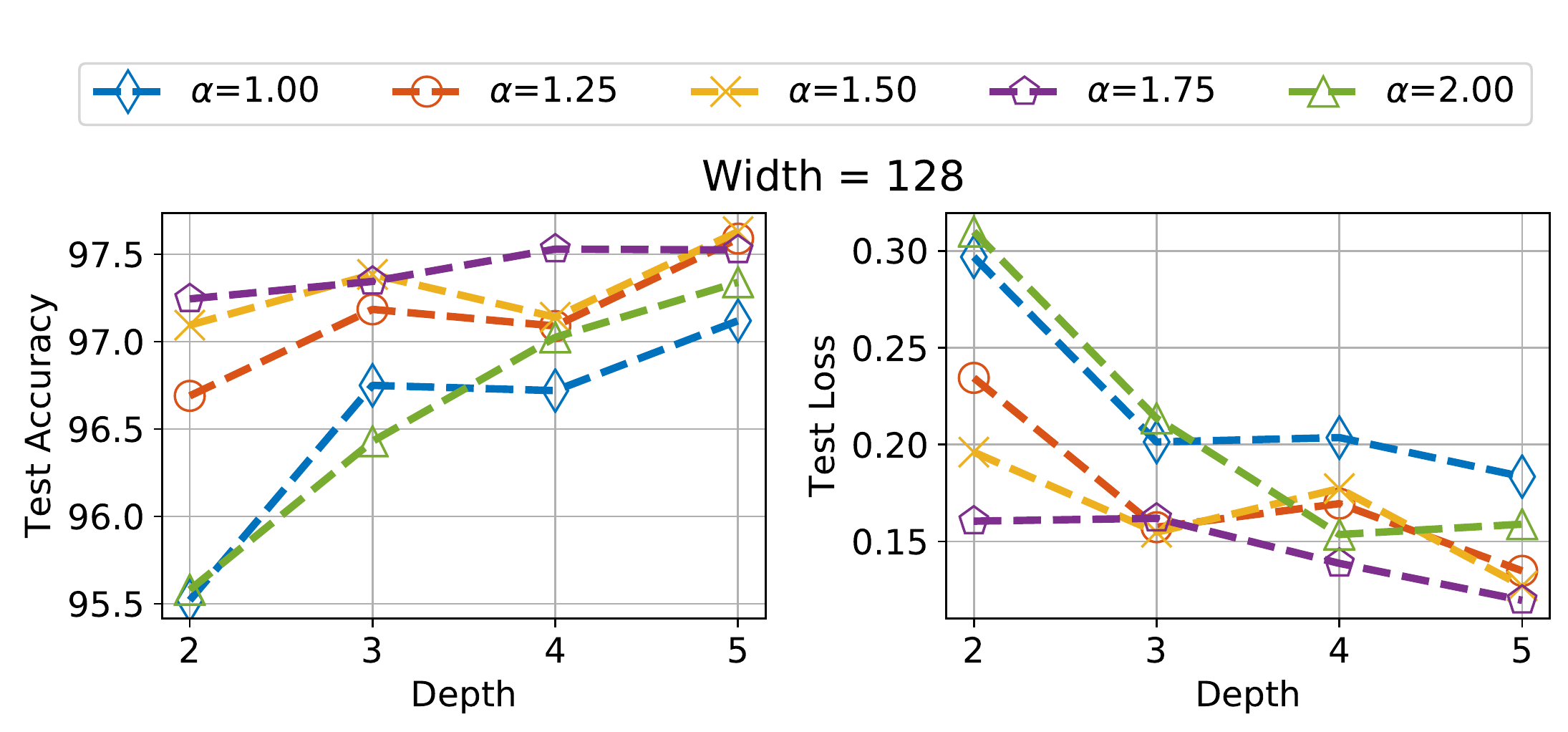}\\
    \includegraphics[width=0.99\columnwidth]{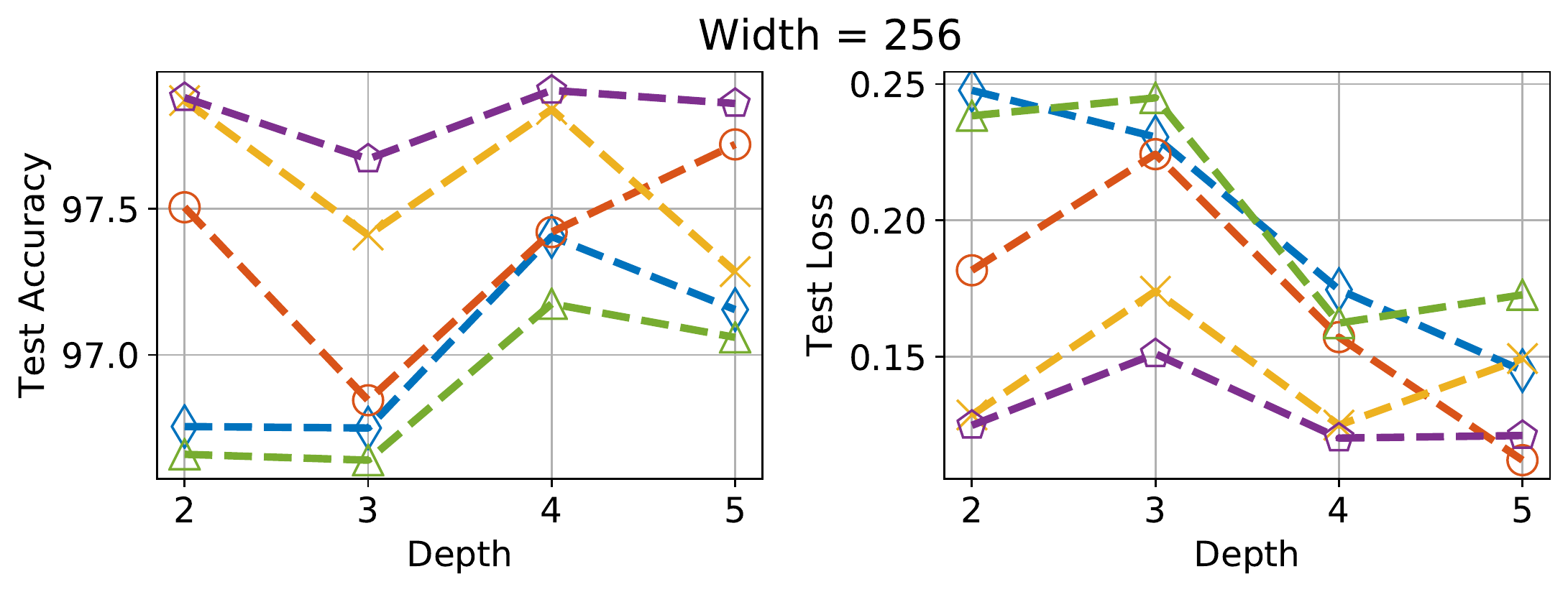}
    \vspace{-10pt}
    \caption{Neural network results on MNIST (test).}
    \label{fig:fcn_mnist2}
\end{figure}

\begin{figure}[t]

    \centering
    \includegraphics[width=0.99\columnwidth]{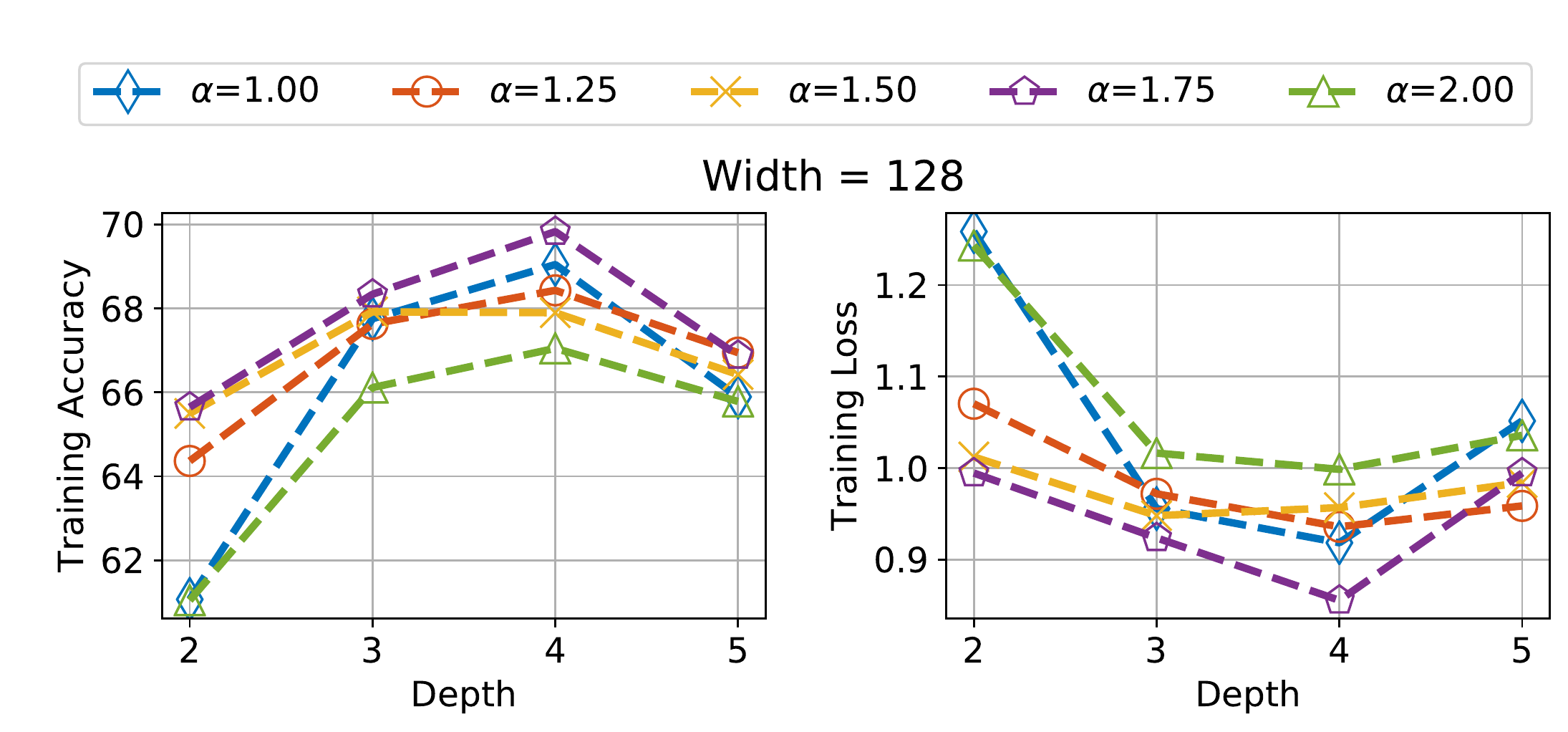}\\
    \includegraphics[width=0.99\columnwidth]{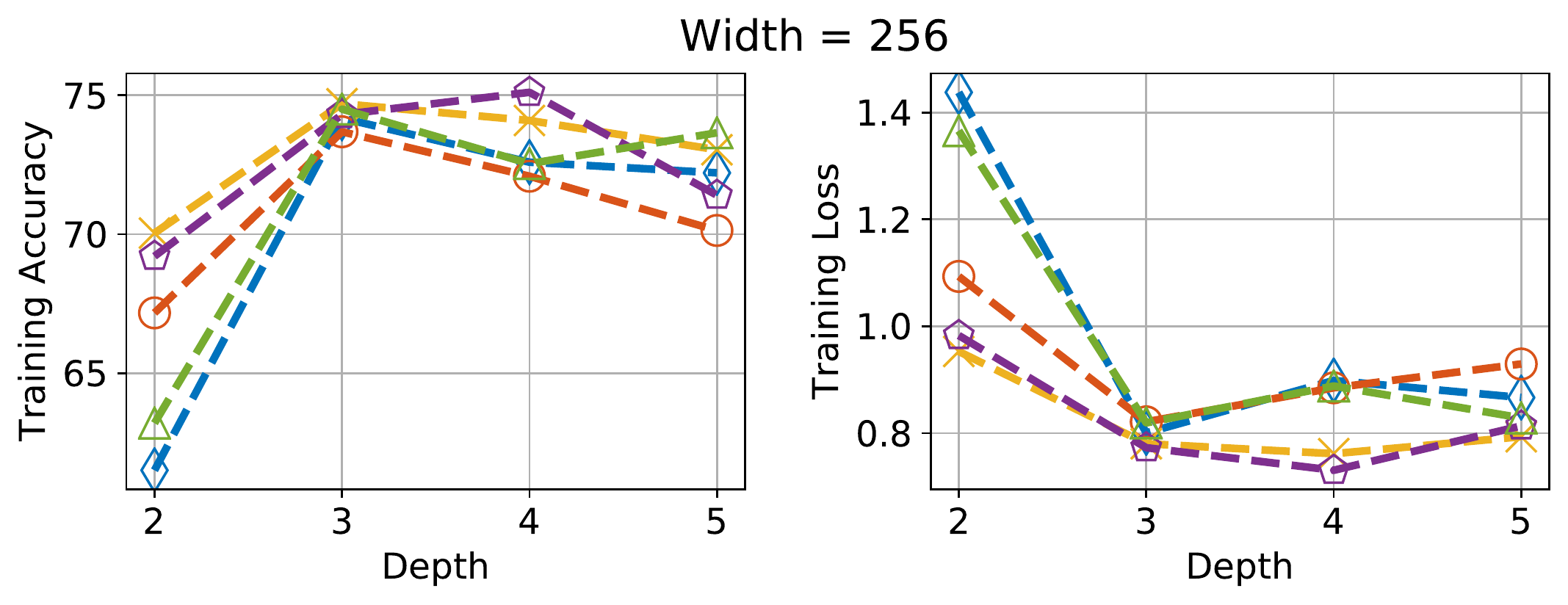}
    \vspace{-10pt}
    \caption{Neural network results on CIFAR10 (training).}
    \label{fig:fcn_cifar}
\end{figure}

\begin{figure}[t]

    \centering
    \includegraphics[width=0.99\columnwidth]{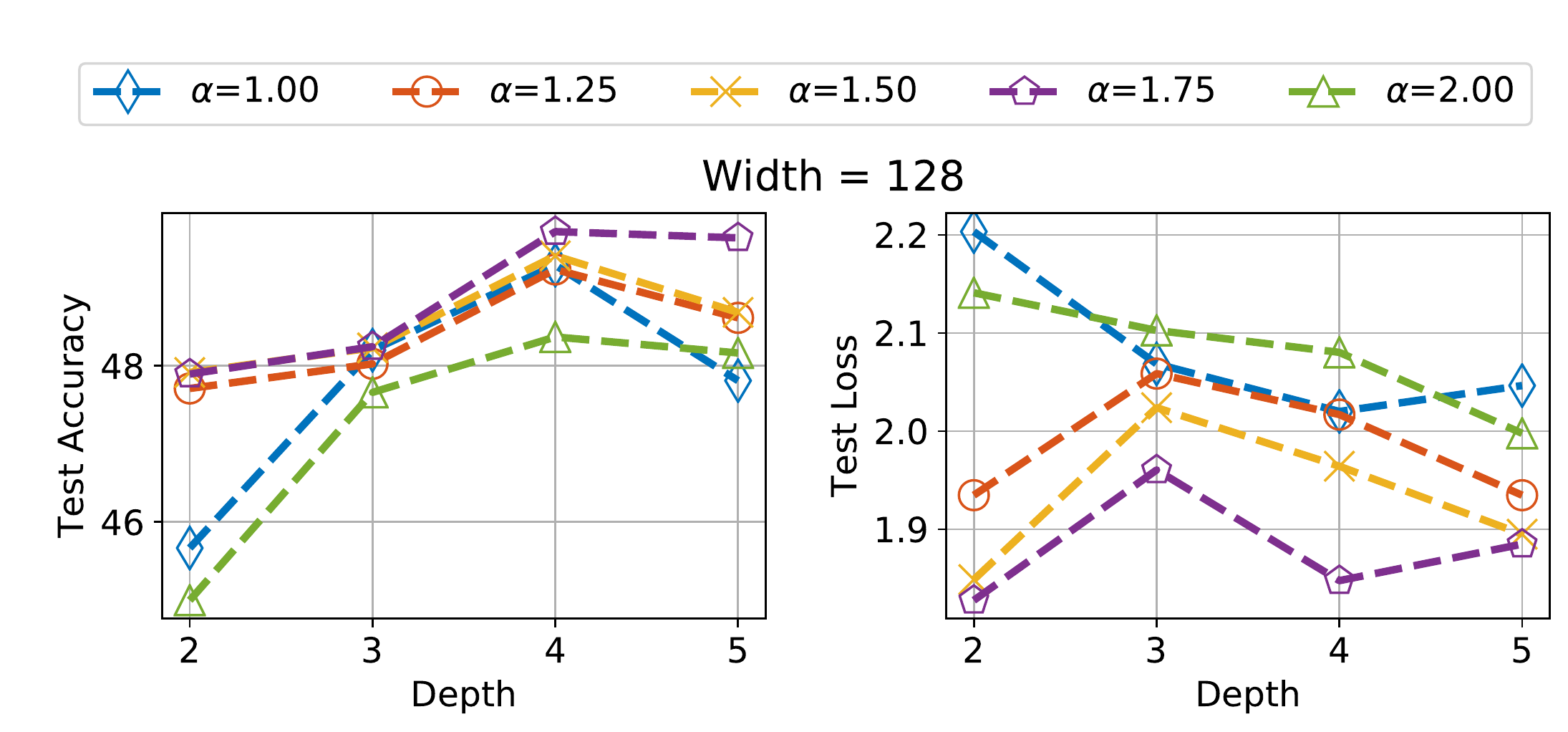}\\
    \includegraphics[width=0.99\columnwidth]{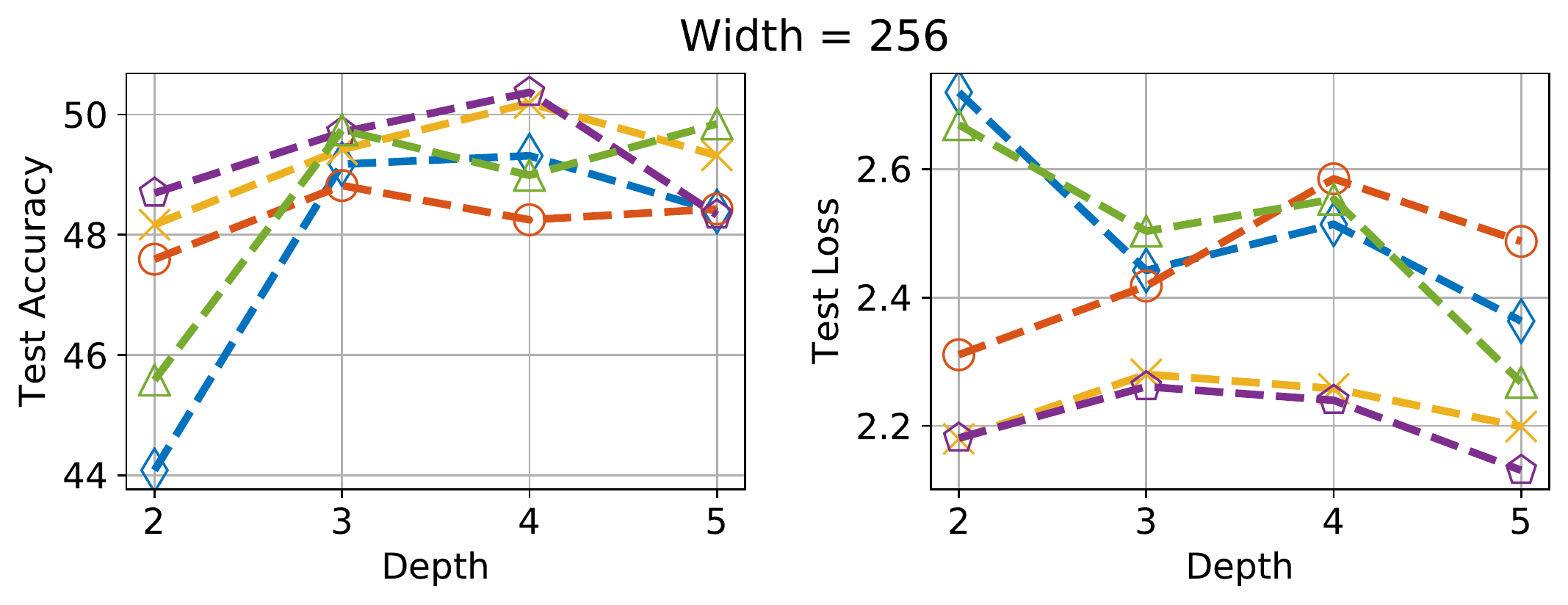}
    \vspace{-10pt}
    \caption{Neural network results on CIFAR10 (test).}
    \label{fig:fcn_cifar2}
\end{figure}

In these experiments, we set $\eta = 0.1$, $\gamma=0.1$ for MNIST, and $\gamma=0.9$ for CIFAR10. We run the algorithms for $K=10000$ iterations \footnote{Since the scale of the gradient noise is proportional to $ (\gamma/\beta)^{\frac1{\alpha}} $ (see \eqref{eqn:eulerv2}), in this setup, a fixed $\gamma$ implicitly determines $\beta$. }. We measure the accuracy and the loss at every 100th iteration and we report the average of the last two measurements. Figures~\ref{fig:fcn_mnist} and \ref{fig:fcn_mnist2} show the results obtained on the MNIST dataset. We observe that, in most of the cases, setting $\alpha=1.75$ yields a better performance in terms both training and testing accuracies/losses. This difference becomes more visible when the width is set to $256$: the accuracy difference between the algorithms reaches $\approx 2\%$. We obtain a similar result on the CIFAR10 dataset, as illustrated in Figures~\ref{fig:fcn_cifar} and \ref{fig:fcn_cifar2}. In most of the cases $\alpha=1.75$ performs better, with the maximum accuracy difference being $\approx 4.5\%$, implying the gradient noise can be approximated by an $\sas$ random variable.

We observed a similar behavior when the width was set to $64$. However, when we set the width to $32$ we did not perceive a significant difference in terms of the performance of the algorithms. 
On the other hand, when the width was set to $512$, $\alpha=2$ resulted in a slightly better performance, which would be an indication that the Gaussian approximation is closer. The corresponding figures are provided in the supplementary document.

\section{Conclusion and Future Directions}

We considered the continuous-time variant of SGDm, known as the underdamped Langevin dynamics (ULD), and developed theory for the case where the gradient noise can be well-approximated by a heavy-tailed $\alpha$-stable random vector. As opposed to na\"{i}vely replacing the driving stochastic force in ULD, which correspondonds to running SGDm with heavy-tailed gradient noise, the dynamics that we developed exactly target the Boltzmann-Gibbs distribution, and hence do not introduce an implicit bias.  We further established the weak convergence of the Euler-Maruyama discretization and illustrated interesting connections between the discretized algorithm and existing approaches commonly used in practice. We supported our theory with experiments on a synthetic setting and fully connected neural networks.

Our framework opens up interesting future directions. Our current modeling strategy requires a state-independent, isotropic noise assumption, which would not accurately reflect the reality. While anisotropic noise can be incorporated to our framework by using the approach of \citet{SFHMC}, state-dependent noise introduces challenging technical difficulties. Similarly, it has been illustrated that the tail-index $\alpha$ can depend on the state and different components of the noise can have a different $\alpha$ \cite{csimcsekli2019heavy}. Incorporating such state dependencies would be an important direction of future research. {Finally, it has been shown that the heavy-tailed perturbations yield shorter escape times \cite{nguyen2019first} in the overdamped dynamics, and extending such results to the underdamped case is still an open problem.}

\section*{Acknowledgments}

We thank Jingzhao Zhang for fruitful discussions. The contribution of Umut \c{S}im\c{s}ekli to this work is partly supported by the French National Research Agency (ANR) as a part of the FBIMATRIX (ANR-16-CE23-0014) project, and by the industrial chair Data science \& Artificial Intelligence from T\'{e}l\'{e}com Paris. Lingjiong Zhu is grateful to the support from Simons Foundation Collaboration Grant. Mert G\"{u}rb\"{u}zbalaban acknowledges support from the grants NSF DMS-1723085 and NSF CCF-1814888. 

\bibliography{langevin}
\bibliographystyle{icml2020}

\newpage

\onecolumn

\icmltitle{Fractional Underdamped Langevin Dynamics: \\Retargeting SGD with Momentum under Heavy-Tailed Gradient Noise\\ {\normalsize SUPPLEMENTARY DOCUMENT}}

\section{Proof of Theorem \ref{thm:invariant}}

\begin{proof}%
Let $q(\xb,\vb,t)$ denote the probability density of $(\xb_{t},\vb_{t})$. 
Then it satisfies the fractional Fokker-Planck equation (see Proposition 1 and Section 7 in \cite{SLDYL}):
\begin{align*}
\partial_{t}q(\xb,\vb,t)
&=\gamma\sum_{i=1}^{d}\frac{\partial[(c(\vb,\alpha))_{i}q(\xb,\vb,t)]}{\partial v_{i}}
+\sum_{i=1}^{d}\frac{\partial[\partial_{x_{i}}f(\xb)q(\xb,\vb,t)]}{\partial v_{i}}
-\frac{\gamma}{\beta}\sum_{i=1}^{d}\mathcal{D}^{\alpha}_{v_{i}}q(\xb,\vb,t)
\\
&\qquad\qquad\qquad
-\sum_{i=1}^{d}\frac{\partial[(\partial_{v_{i}}g(\vb)q(\xb,\vb,t)]}{\partial x_{i}}.
\end{align*}
We can compute that
\begin{align}
&\gamma\sum_{i=1}^{d}\frac{\partial[(c(\vb,\alpha))_{i}\phi(\xb)\psi(\vb)]}{\partial v_{i}}
+\sum_{i=1}^{d}\frac{\partial[\partial_{x_{i}}f(\xb)\phi(\xb)\psi(\vb)]}{\partial v_{i}}
-\frac{\gamma}{\beta}\sum_{i=1}^{d}\mathcal{D}^{\alpha}_{v_{i}}\phi(\xb)\psi(\vb)
-\sum_{i=1}^{d}\frac{\partial[(\partial_{v_{i}}g(\vb)\phi(\xb)\psi(\vb)]}{\partial x_{i}}
\nonumber
\\
&\quad =\frac{\gamma}{\beta}\phi(\xb)\left[\beta\sum_{i=1}^{d}\frac{\partial[(c(\vb,\alpha))_{i}\psi(\vb)]}{\partial v_{i}}
-\sum_{i=1}^{d}\mathcal{D}^{\alpha}_{v_{i}}\psi(\vb)\right]
+\sum_{i=1}^{d}\frac{\partial[\partial_{x_{i}}f(\xb)\phi(\xb)\psi(\vb)]}{\partial v_{i}}
-\sum_{i=1}^{d}\frac{\partial[\partial_{v_{i}}g(\vb)\phi(\xb)\psi(\vb)]}{\partial x_{i}}.\label{by:1}
\end{align}
Furthermore, we can compute that
\begin{align}
\beta\sum_{i=1}^{d}\frac{\partial[(c(\vb,\alpha))_{i}\psi(\vb)]}{\partial v_{i}}
-\sum_{i=1}^{d}\mathcal{D}^{\alpha}_{v_{i}}\psi(\vb)
\nonumber
&=\sum_{i=1}^{d}\frac{\partial}{\partial v_{i}}
\left[\frac{\mathcal{D}_{v_{i}}^{\alpha-2}(\psi(\vb)\partial_{v_{i}}\beta g(\vb))}{\psi(\vb)}\psi(\vb)\right]
-\sum_{i=1}^{d}\mathcal{D}^{\alpha}_{v_{i}}\psi(\vb)
\nonumber
\\
&=-\sum_{i=1}^{d}\frac{\partial}{\partial v_{i}}
\left[\frac{\mathcal{D}_{v_{i}}^{\alpha-2}(\partial_{v_{i}}\psi(\vb))}{\psi(\vb)}\psi(\vb)\right]
-\sum_{i=1}^{d}\mathcal{D}^{\alpha}_{v_{i}}\psi(\vb)
\nonumber
\\
&=-\sum_{i=1}^{d}\frac{\partial^{2}}{\partial v_{i}^{2}}
\mathcal{D}_{v_{i}}^{\alpha-2}\psi(\vb)
-\sum_{i=1}^{d}\mathcal{D}^{\alpha}_{v_{i}}\psi(\vb)
\nonumber
\\
&=\sum_{i=1}^{d}\mathcal{D}_{v_{i}}^{2}
\mathcal{D}_{v_{i}}^{\alpha-2}\psi(\vb)
-\sum_{i=1}^{d}\mathcal{D}^{\alpha}_{v_{i}}\psi(\vb)=0,\label{by:2}
\end{align}
where we used the property $\mathcal{D}^{2}u(x)=-\frac{\partial^{2}}{\partial x^{2}}u(x)$ (Proposition 1 in \cite{FLMC})
and the semi-group property of the Riesz derivative $\mathcal{D}^{a}\mathcal{D}^{b}u(x)=\mathcal{D}^{a+b}u(x)$.

Finally, we can compute that
\begin{align}
&\sum_{i=1}^{d}\frac{\partial[\partial_{x_{i}}f(\xb)\phi(\xb)\psi(\vb)]}{\partial v_{i}}
-\sum_{i=1}^{d}\frac{\partial[\partial_{v_{i}}g(\vb)\phi(\xb)\psi(\vb)]}{\partial x_{i}}
\nonumber
\\
&=\sum_{i=1}^{d}\partial_{x_{i}}f(\xb)\phi(\xb)\partial_{v_{i}}\psi(\vb)
-\sum_{i=1}^{d}\partial_{v_{i}}g(\vb)\psi(\vb)\partial_{x_{i}}\phi(\xb)
\nonumber
\\
&=-\beta\sum_{i=1}^{d}\frac{\partial f(\xb)}{\partial_{x_{i}}}\phi(\xb)\frac{\partial g(\vb)}{\partial v_{i}}\psi(\vb)
+\beta\sum_{i=1}^{d}\frac{\partial g(\vb)}{\partial_{v_{i}}}\psi(\vb)\frac{\partial f(\xb)}{\partial x_{i}}\phi(\xb)=0.\label{by:3}
\end{align}
Therefore, it follows from \eqref{by:1}, \eqref{by:2} and \eqref{by:3} that we have
\begin{equation*}
\gamma\sum_{i=1}^{d}\frac{\partial[(c(\vb,\alpha))_{i}\phi(\xb)\psi(\vb)]}{\partial v_{i}}
+\sum_{i=1}^{d}\frac{\partial[\partial_{x_{i}}f(\xb)\phi(\xb)\psi(\vb)]}{\partial v_{i}}
-\frac{\gamma}{\beta}\sum_{i=1}^{d}\mathcal{D}^{\alpha}_{v_{i}}\phi(\xb)\psi(\vb)
-\sum_{i=1}^{d}\frac{\partial[(\partial_{v_{i}}g(\vb)\phi(\xb)\psi(\vb)]}{\partial x_{i}}=0.
\end{equation*}
Hence we conclude that  $\pi(d\xb,d\vb)=\frac{e^{-\beta(f(\xb)+g(\vb))}d\xb d\vb}{\int_{\mathbb{R}^{d}\times\mathbb{R}^{d}}e^{-\beta(f(\xb')+g(\vb'))}d\xb'd\vb'}$
is an invariant probability measure.
The proof is complete.
\end{proof}

\section{Proof of Theorem \ref{prop:formula}}

\begin{proof}%
We can compute that
\begin{equation}
(c(\vb,\alpha))_{i}=\frac{\mathcal{D}_{v_{i}}^{\alpha-2}(v_{i}e^{-\frac{1}{2}\Vert \vb\Vert^{2}})}{e^{-\frac{1}{2}\Vert \vb\Vert^{2}}}
=e^{\frac{1}{2}v_{i}^{2}}\mathcal{D}_{v_{i}}^{\alpha-2}\left(v_{i}e^{-\frac{1}{2}v_{i}^{2}}\right),
\end{equation}
for every $1\leq i\leq d$.

Recall the definition of Fourier transform and its inverse:
\begin{equation}
\mathcal{F}\{f(x)\}(\omega)
=\frac{1}{\sqrt{2\pi}}\int_{-\infty}^{\infty}e^{-ix\omega}f(x)dx,
\qquad
\mathcal{F}^{-1}\{f(\omega)\}(x)
=\frac{1}{\sqrt{2\pi}}\int_{-\infty}^{\infty}e^{ix\omega}f(\omega)d\omega.
\end{equation}
Notice that the Fourier transform of $e^{-\frac{1}{2}x^{2}}$ is itself,
i.e. $\mathcal{F}\{e^{-\frac{1}{2}x^{2}}\}(\omega)=e^{-\frac{1}{2}\omega^{2}}$,
and moreover, $\mathcal{F}\{x^{n}f(x)\}(\omega)=i^{n}\frac{d^{n}}{d\omega^{n}}\{\mathcal{F}\{f(x)\}(\omega)\}$,
and therefore,
\begin{equation}
\mathcal{F}\left\{xe^{-\frac{1}{2}x^{2}}\right\}(\omega)
=-i\omega e^{-\frac{1}{2}\omega^{2}}.
\end{equation}
Hence,
\begin{equation}
\mathcal{D}_{x}^{\alpha-2}\left(xe^{-\frac{1}{2}x^{2}}\right)
=\mathcal{F}^{-1}\left\{-i\omega|\omega|^{\alpha-2}e^{-\frac{1}{2}\omega^{2}}\right\}(x)
=\frac{-i}{\sqrt{2\pi}}\int_{-\infty}^{\infty}\omega|\omega|^{\alpha-2}e^{-\frac{1}{2}\omega^{2}+i\omega x}d\omega.
\end{equation}
Furthermore, we can compute that
\begin{align*}
\frac{-i}{\sqrt{2\pi}}\int_{-\infty}^{\infty}\omega|\omega|^{\alpha-2}e^{-\frac{1}{2}\omega^{2}+i\omega x}d\omega
&=\frac{-i}{\sqrt{2\pi}}\int_{0}^{\infty}\omega^{\alpha-1}e^{-\frac{1}{2}\omega^{2}+i\omega x}d\omega
+\frac{-i}{\sqrt{2\pi}}\int_{-\infty}^{0}\omega(-\omega)^{\alpha-2}e^{-\frac{1}{2}\omega^{2}+i\omega x}d\omega
\\
&=\frac{-i}{\sqrt{2\pi}}\int_{0}^{\infty}\omega^{\alpha-1}e^{-\frac{1}{2}\omega^{2}+i\omega x}d\omega
+\frac{i}{\sqrt{2\pi}}\int_{0}^{\infty}\omega^{\alpha-1}e^{-\frac{1}{2}\omega^{2}-i\omega x}d\omega
\\
&=\sqrt{\frac{2}{\pi}}\int_{0}^{\infty}\omega^{\alpha-1}\sin(\omega x)e^{-\frac{1}{2}\omega^{2}}d\omega.
\end{align*}
By the Taylor expansion of sine function, we get
\begin{align*}
\sqrt{\frac{2}{\pi}}\int_{0}^{\infty}\omega^{\alpha-1}\sin(\omega x)e^{-\frac{1}{2}\omega^{2}}d\omega
&=\sqrt{\frac{2}{\pi}}\int_{0}^{\infty}\omega^{\alpha-1}\sum_{k=0}^{\infty}\frac{(-1)^{k}(\omega x)^{2k+1}}{(2k+1)!}
e^{-\frac{1}{2}\omega^{2}}d\omega
\\
&=\sqrt{\frac{2}{\pi}}\sum_{k=0}^{\infty}\frac{(-1)^{k}x^{2k+1}}{(2k+1)!}\int_{0}^{\infty}\omega^{2k+\alpha}
e^{-\frac{1}{2}\omega^{2}}d\omega
\\
&=\sqrt{\frac{2}{\pi}}\sum_{k=0}^{\infty}\frac{(-1)^{k}x^{2k+1}}{(2k+1)!}2^{\frac{2k+\alpha-1}{2}}
\Gamma\left(\frac{2k+\alpha+1}{2}\right),
\end{align*}
where we used the identity $\int_{0}^{\infty}x^{a}e^{-\frac{1}{2}x^{2}}dx=2^{\frac{a-1}{2}}\Gamma(\frac{a+1}{2})$,
for any given $a>-1$.
Moreover, for any given $x,y>0$, we have the identity:
\begin{equation}
\sum_{k=0}^{\infty}\frac{(-1)^{k}x^{k}}{(2k+1)!}\Gamma(k+y)
=\Gamma(y)_{1}F_{1}\left(y;\frac{3}{2};-\frac{x}{4}\right),
\end{equation}
where $_{1}F_{1}$ is the Kummer confluent hypergeometric function.
Therefore, we conclude that
\begin{align*}
\sqrt{\frac{2}{\pi}}\sum_{k=0}^{\infty}\frac{(-1)^{k}x^{2k+1}}{(2k+1)!}2^{\frac{2k+\alpha-1}{2}}
\Gamma\left(\frac{2k+\alpha+1}{2}\right)
&=\sqrt{\frac{2}{\pi}}2^{\frac{\alpha-1}{2}}x\sum_{k=0}^{\infty}\frac{(-1)^{k}(2x^{2})^{k}}{(2k+1)!}
\Gamma\left(k+\frac{\alpha+1}{2}\right)
\\
&=\frac{2^{\frac{\alpha}{2}}x}{\sqrt{\pi}}\Gamma\left(\frac{\alpha+1}{2}\right)
\cdot_{1}F_{1}\left(\frac{\alpha+1}{2};\frac{3}{2};-\frac{x^{2}}{2}\right).
\end{align*}
Hence, we get for every $1\leq i\leq d$,
\begin{equation}
(c(\vb,\alpha))_{i}
=\frac{2^{\frac{\alpha}{2}}v_{i}e^{\frac{1}{2}v_{i}^{2}}}{\sqrt{\pi}}\Gamma\left(\frac{\alpha+1}{2}\right)
\cdot_{1}F_{1}\left(\frac{\alpha+1}{2};\frac{3}{2};-\frac{v_{i}^{2}}{2}\right).
\end{equation}
By the identity $e^{x}\cdot_{1}F_{1}(a;b;-x)=_{1}F_{1}(b-a;b;x)$, we get
\begin{equation}
(c(\vb,\alpha))_{i}
=\frac{2^{\frac{\alpha}{2}}v_{i}}{\sqrt{\pi}}\Gamma\left(\frac{\alpha+1}{2}\right)
\cdot_{1}F_{1}\left(\frac{2-\alpha}{2};\frac{3}{2};\frac{v_{i}^{2}}{2}\right).
\end{equation}
In particular, when $\alpha=2$, by applying the identity
\begin{equation}
\sum_{k=0}^{\infty}\frac{(-1)^{k}x^{k}}{(2k+1)!}\Gamma\left(k+\frac{3}{2}\right)
=\frac{\sqrt{\pi}}{2}e^{-x/4},
\end{equation}
we get
\begin{equation*}
\sqrt{\frac{2}{\pi}}\sum_{k=0}^{\infty}\frac{(-1)^{k}x^{2k+1}}{(2k+1)!}2^{\frac{2k+1}{2}}
\Gamma\left(\frac{2k+3}{2}\right)
=\sqrt{\frac{2}{\pi}}2^{\frac{1}{2}}x\sum_{k=0}^{\infty}\frac{(-1)^{k}(2x^{2})^{k}}{(2k+1)!}
\Gamma\left(k+\frac{3}{2}\right)=xe^{-\frac{x^{2}}{2}}.
\end{equation*}
The proof is complete.
\end{proof}

\section{Proof of Theorem \ref{prop:v}}
\begin{proof}%
Let $\psi_{\alpha}(x)=e^{-g_{\alpha}(x)}$ be the probability density
function of the symmetric $\alpha$-stable distribution $\mathcal{S}\alpha\mathcal{S}(\frac{1}{\alpha^{1/\alpha}})$
such that
\begin{equation}
\mathcal{F}\{\psi_{\alpha}(x)\}(\omega)
=\frac{1}{\sqrt{2\pi}}\int_{-\infty}^{\infty}e^{-i\omega x}\psi_{\alpha}(x)dx
=\frac{1}{\sqrt{2\pi}}e^{-\frac{1}{\alpha}|\omega|^{\alpha}}.
\end{equation}
Therefore, we get
\begin{align*}
\mathcal{D}_{x}^{\alpha-2}(\psi_{\alpha}(x)\partial_{x}g_{\alpha}(x))
&=-\mathcal{D}_{x}^{\alpha-2}(\partial_{x}\psi_{\alpha}(x))
\\
&=-\mathcal{F}^{-1}\left\{|\omega|^{\alpha-2}\mathcal{F}\left\{\partial_{x}\psi_{\alpha}(x)\right\}(\omega)\right\}(x)
\\
&=-\mathcal{F}^{-1}\left\{|\omega|^{\alpha-2}(i\omega)\mathcal{F}\left\{\psi_{\alpha}(x)\right\}(\omega)\right\}(x)
\\
&=\frac{-i}{\sqrt{2\pi}}\mathcal{F}^{-1}\left\{|\omega|^{\alpha-2}\omega e^{-\frac{1}{\alpha}|\omega|^{\alpha}}\right\}(x)
\\
&=\frac{i}{\sqrt{2\pi}}\mathcal{F}^{-1}\left\{\partial_{\omega}e^{-\frac{1}{\alpha}|\omega|^{\alpha}}\right\}(x)
\\
&=\frac{i}{\sqrt{2\pi}}(-ix)\mathcal{F}^{-1}\left\{e^{-\frac{1}{\alpha}|\omega|^{\alpha}}\right\}(x)
\\
&=x\psi_{\alpha}(x).
\end{align*}
Hence, we conclude that
\begin{equation}
\frac{\mathcal{D}_{x}^{\alpha-2}(\psi_{\alpha}(x)\partial_{x}g_{\alpha}(x))}{\psi_{\alpha}(x)}=x,
\end{equation}
and it follows that
\begin{equation}
(c(\vb,\alpha))_{i}=v_{i},\qquad 1\leq i\leq d.
\end{equation}
The proof is complete.
\end{proof}

\section{Proof of Proposition \ref{prop:lipschitz}}
\begin{proof}%
It is straightforward to verify that the result holds for the cases $\alpha=1$ and $\alpha=2$. Assume $\alpha \in (0,1)$ or $\alpha \in (1,2)$. Let $X$ be the unit symmetric $\alpha$-stable random variable defined by its characteristic function
$$\phi_X(t) := \mathbb{E}(e^{itX}) = e^{-|t|^\alpha}.$$
By taking inverse Fourier transformation, its density
$\psi_\alpha(x)=e^{-g_{\alpha}(x)}$ can be expressed as
\begin{equation*}
    \psi_\alpha(x) := \frac{1}{2\pi} \int_{-\infty}^\infty \phi_X(t) e^{-itx}dt.
\end{equation*}
Writing $e^{-itx} = \cos(tx) -i \sin(tx)$, we compute
\begin{equation}\label{eq-int-density}
 \psi_\alpha(x) = \frac{1}{2\pi} \int_{-\infty}^\infty e^{-|t|^\alpha} \left[\cos(tx) -i \sin(tx)\right] dt
 =\frac{1}{\pi} \int_0^\infty  e^{-t^\alpha} \cos(tx) dt \,,
\end{equation}
where we used the fact that $\phi_X(t)$ and $\cos(tx)$ are even functions of $t$, whereas $\sin(tx)$ is an odd function of $t$. If we define, 
$$g_\alpha(x) = -\log(\psi_\alpha(x)),$$
then
\begin{equation}
 g'_{\alpha}(x) = \frac{\psi'_\alpha(x)}{\psi_\alpha(x)}\,,
\end{equation}
where the superscript $'$ denotes derivative with respect to $x$. Similarly, 
\begin{equation}\label{def-g-alpha-diff2}
 g''_{\alpha}(x) = \frac{\psi''_\alpha(x)}{\psi_\alpha(x)} - \left(\frac{\psi'_\alpha(x)}{\psi_\alpha(x)}\right)^2\,.
\end{equation}
If $g''_{\alpha}(x)$ is uniformly bounded over $x\in\mathbb{R}$, it can be seen that the map $v\mapsto  g'_\alpha(v)$ will be Lipschitz. Therefore, it suffices to show that $x\mapsto g''_{\alpha}(x)$ is a bounded function on the real line. Note that the function $\psi_\alpha(x)$ is infinitely many differentiable, and the integral \eqref{eq-int-density} is absolutely convergent. Therefore, we can differentiate both sides of \eqref{eq-int-density} with respect to $x$ to obtain
\begin{align*}
&\psi'_\alpha(x):= \frac{1}{\pi} \int_0^\infty -t  e^{-t^\alpha} \sin(tx) dt\,,
\\
&\psi''_\alpha(x):= \frac{1}{\pi} \int_0^\infty -t^2  e^{-t^\alpha} \cos(tx) dt\,. 
\end{align*}
In particular, since $|\cos(tx)|\leq 1$ and $|\sin(tx)|\leq 1$ this implies that
\begin{align*}
&\psi'_\alpha(x) \leq M_1(\alpha):= \frac{1}{\pi} \int_0^\infty t  e^{-t^\alpha}  dt <\infty\,, 
\\
&\psi''_\alpha(x) \leq M_2(\alpha):=\frac{1}{\pi} \int_0^\infty t^2  e^{-t^\alpha}dt <\infty\,. 
\end{align*}
It is also well-known that a symmetric $\alpha$ stable random variable has a decay in its density satisfying $\psi_\alpha(x)\sim \frac{1}{|x|^{1+\alpha}}$ when $|x|$ is large. In fact, \citet{wintner1941} derived a large-$x$ expansion for $\psi_\alpha(x)$ when $0<\alpha<1$ and $x>0$. This expansion is equivalent to 
\begin{align*}
\psi_\alpha(x) &= \frac{1}{\pi} \sum_{n=1}^\infty \frac{(-1)^{n+1}}{n!}
\frac{\Gamma(1+\alpha n)}{x^{\alpha n +1}}\sin\left(\frac{\pi \alpha n}{2}\right) \\
&= \frac{1}{\pi}
\bigg( \frac{\Gamma(1+\alpha)}{x^{\alpha+1}} \sin(\pi \alpha/2)
- 
\frac{\Gamma(1+2\alpha)}{2x^{2\alpha+1}} \sin(\pi \alpha)
+
\frac{\Gamma(1+3\alpha)}{6x^{3\alpha+1}} \sin(\pi 3\alpha/2) + \cdots \bigg)\,,
\end{align*}
(see eqn. (11) from \cite{Montroll1984}) where it can be seen from the Stirling's approximation of the gamma function and the ratio test that the series converges absolutely. A similar absolutely convergent series sum (with exactly the same leading term) is also available in the literature for $\alpha \in (1,2)$ which says that
$$\psi_\alpha(x) = \frac{1}{\pi}
 \frac{\Gamma(1+\alpha)}{x^{\alpha+1}} \sin\left(\frac{\pi \alpha}{2}\right) + O\left(\frac{1}{x^{2\alpha}}\right)
$$
(see eqn. (3.58) from
\cite{MONTROLL_book_chapter}).
By differentiating the series sum for $\psi_\alpha(x)$ with respect to $x$, we can express $\psi'_\alpha(x)$ and $\psi''_\alpha(x)$ as a series sum. After a straightforward computation, we obtain

$$ \frac{\psi''_\alpha(x)}{\psi_\alpha(x)} = O\left(\frac{1}{x^2}\right), \quad \left(\frac{\psi'_\alpha(x)}{\psi_\alpha(x)}\right)^2 = O\left(\frac{1}{x^2}\right),$$
which implies from \eqref{def-g-alpha-diff2} that $g''_\alpha(x)\to 0$ as $x\to\infty$. This shows that $g''_\alpha(x)$ is bounded on the interval $[0,\infty)$. On the other hand, $\psi_\alpha(x)$ is an even function and therefore $g''_\alpha(x)$ is an even function satisfying $g''_\alpha(x)=g''_\alpha(-x)$. We conclude that $g''_\alpha(x)$ is bounded on the real line. This completes the proof.
\end{proof}

\section{Proof of Corollary~\ref{cor:weakconv}}

\begin{proof}
By Proposition \ref{prop:lipschitz}, we know that $\nabla G_\alpha$ is Lipschitz and by our hypthesis $\nabla f$ is also Lipschitz and has linear growth. Then the process \eqref{eqn-sde} admits a unique invariant measure (cf.\ \cite{SLDYL} Section 9), which is given by Theorem~\ref{thm:invariant}. The rest of the proof follows from \cite{panloup2008recursive} (Theorem 2).
\end{proof}

 \begin{figure}[t]
    \centering
    \includegraphics[width=0.7\columnwidth]{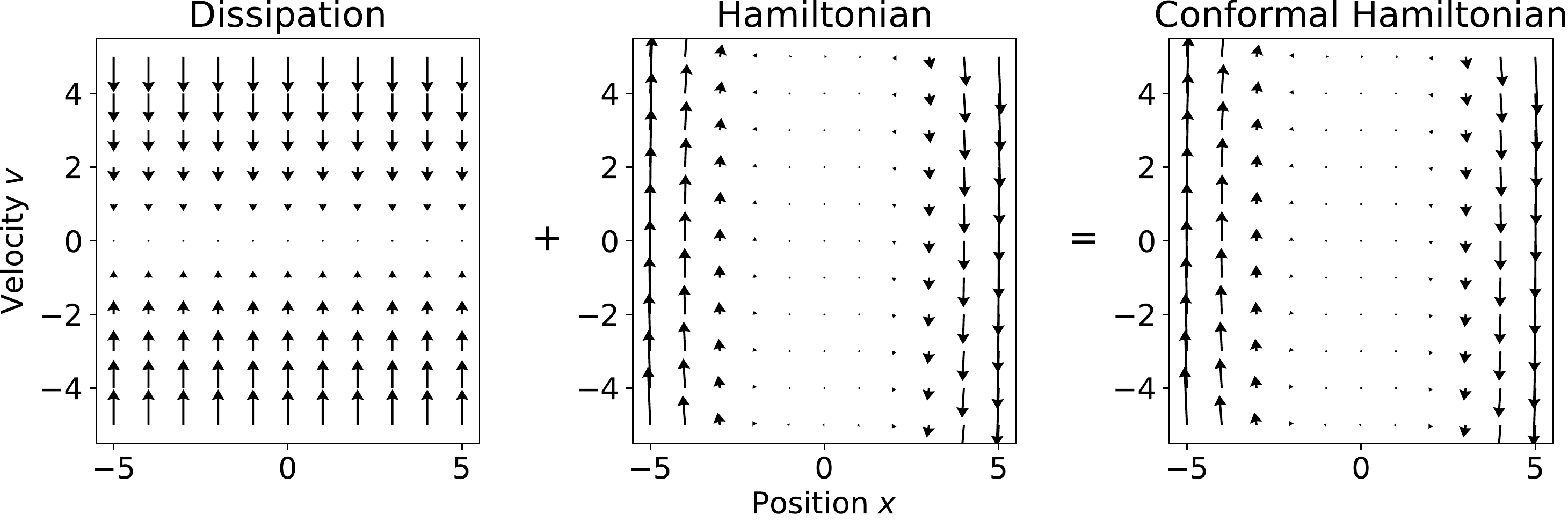} \vspace{5pt} \\
    \includegraphics[width=0.7\columnwidth]{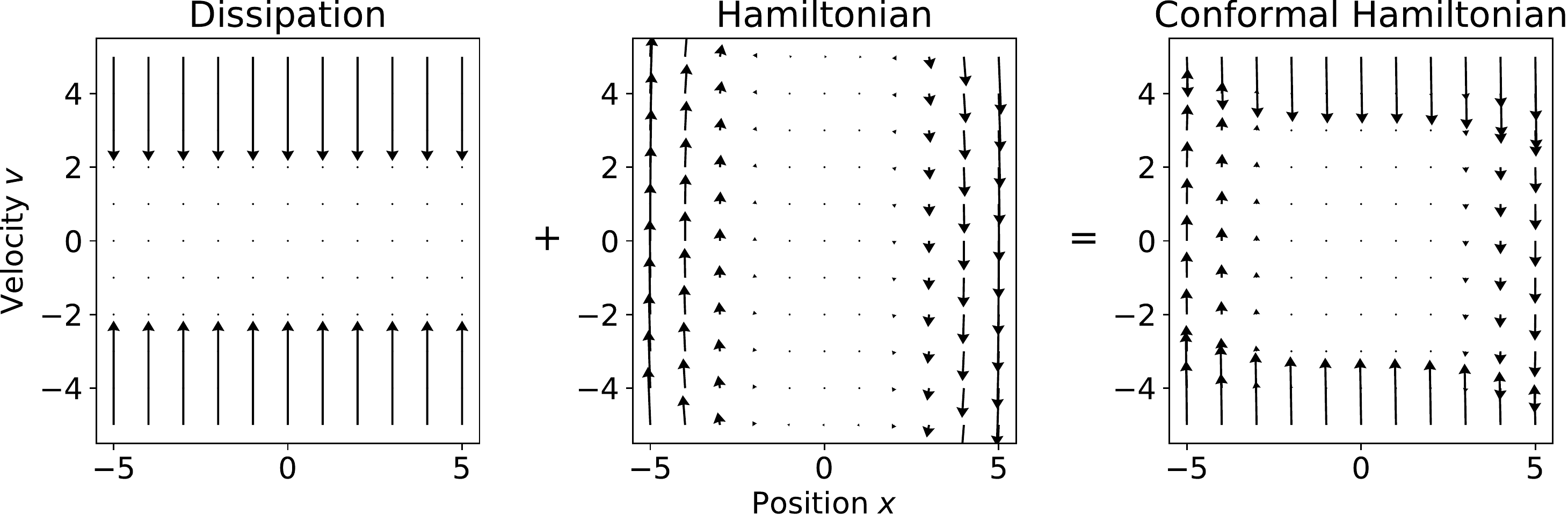} 
    \vspace{-10pt}
    \caption{Conformal Hamiltonian fields with the Gaussian kinetic energy for $f(x) = x^4/4$. Top $\alpha=2$, bottom $\alpha=1.7$.}
    \label{fig:quiver_v1}
\end{figure}

\section{Alternative forms of the drift function $c$ with the Gaussian kinetic energy}

For some special values of $\alpha$, we can get alternative formulas
for $(c(\vb,\alpha))_{i}$, $1\leq i\leq d$.

(1) $\alpha=\frac{3}{2}$. Using the identity 
$_{1}F_{1}(a;2a+1;z)=2^{2a-1}\Gamma(a+\frac{1}{2})e^{\frac{z}{2}}z^{\frac{1}{2}-a}
(I_{a-\frac{1}{2}}(\frac{z}{2})-I_{a+\frac{1}{2}}(\frac{z}{2}))$, 
where $I_{a}(x)$ is the modified Bessel function of the first kind, we get
\begin{equation}
(c(\vb,\alpha))_{i}
=\frac{2^{\frac{1}{4}}v_{i}}{\sqrt{\pi}}\Gamma\left(\frac{5}{4}\right)
\Gamma\left(\frac{3}{4}\right)e^{\frac{v_{i}^{2}}{4}}\left(\frac{v_{i}^{2}}{2}\right)^{\frac{1}{4}}
\left(I_{-\frac{1}{4}}\left(\frac{v_{i}^{2}}{4}\right)-I_{\frac{3}{4}}\left(\frac{v_{i}^{2}}{4}\right)\right),
\end{equation}
for every $1\leq i\leq d$.

(2) $\alpha=\frac{1}{2}$. Using the identity
$_{1}F_{1}(a;2a;z)=2^{2a-1}\Gamma(a+\frac{1}{2})z^{\frac{1}{2}-a}e^{\frac{z}{2}}I_{a-\frac{1}{2}}(\frac{z}{2})$,
we get
\begin{equation}
(c(\vb,\alpha))_{i}
=\frac{2^{\frac{3}{4}}v_{i}}{\sqrt{\pi}}
\Gamma\left(\frac{3}{4}\right)
\Gamma\left(\frac{5}{4}\right)\left(\frac{v_{i}^{2}}{2}\right)^{-\frac{1}{4}}e^{\frac{v_{i}^{2}}{4}}
I_{\frac{1}{4}}\left(\frac{v_{i}^{2}}{4}\right),
\end{equation}
for every $1\leq i\leq d$.

\section{Visual Illustrations}

In order to have a better grasp on the dynamics \eqref{eqn:fuld_v1} in an optimization context, we also investigate its deterministic part (i.e., \eqref{eqn:fuld_v1} without the $\Lm_t$ term) as a \emph{conformal Hamiltonian system} \cite{maddison2018hamiltonian}, where we decompose the overall dynamics into two: the dissipative part $\rmd(\xb_{t},\vb_{t}) = (0, -\gamma c(\vb_{t-},\alpha))\rmd t$ and the Hamiltonian part $\rmd(\xb_{t},\vb_{t})= (\vb_t,-\nabla f(\xb_{t}))\rmd t$, whose combination gives the conformal Hamiltonian. The two parts have different semantics: the Hamiltonian part tries to keep the overall energy of the system ($\nabla f(\xb) + \|\vb\|^2/2$) constant, while the dissipative part tries to reduce this energy, and this competition determines the behavior of the overall system. In Figure~\ref{fig:quiver_v1}, we visualize the conformal Hamiltonians for $f(x) = x^4/4$ for two different values of $\alpha$. This choice of $f$ is known to be problematic for the classical overdamped dynamics \cite{maddison2018hamiltonian,brosse2019tamed}, which can be clearly observed from Figure~\ref{fig:quiver_v1} (top right) as the conformal Hamiltonian field tends to diverge. On the other hand, for $\alpha=1.7$, we observe that the strong dissipation, which was introduced due to tolerate heavy-tailed perturbations, can also compensate for fast-growing $f$.

\begin{figure}[t]
    \centering
    \includegraphics[width=0.7\columnwidth]{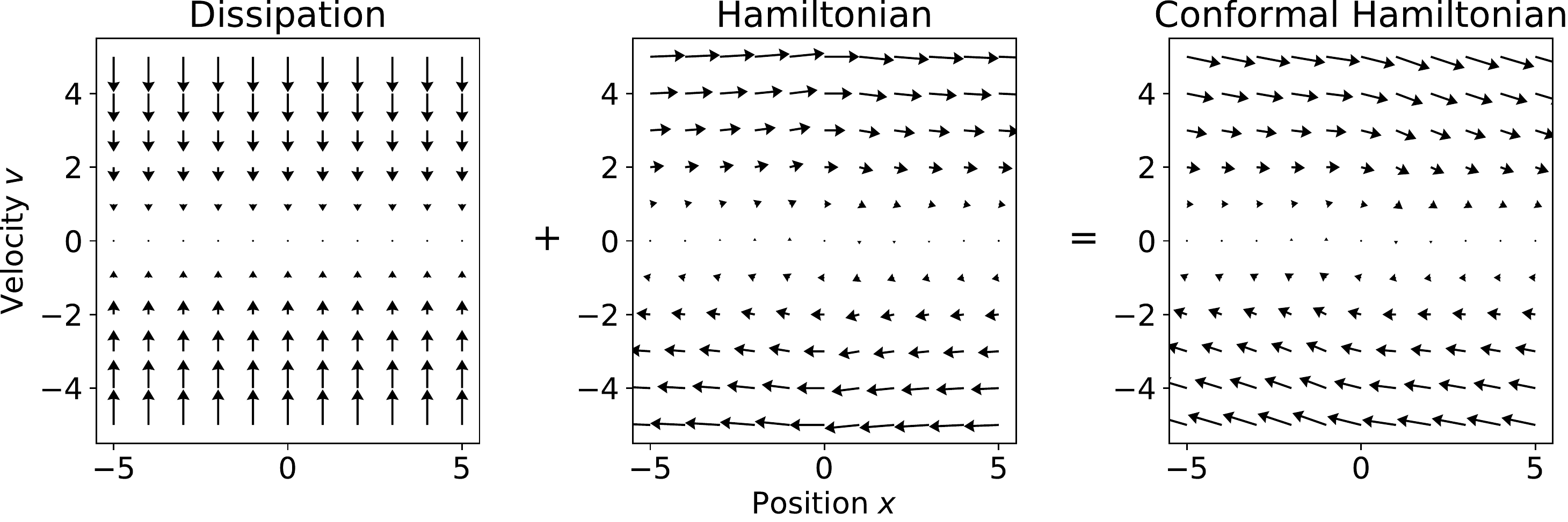} \vspace{5pt} \\
    \includegraphics[width=0.7\columnwidth]{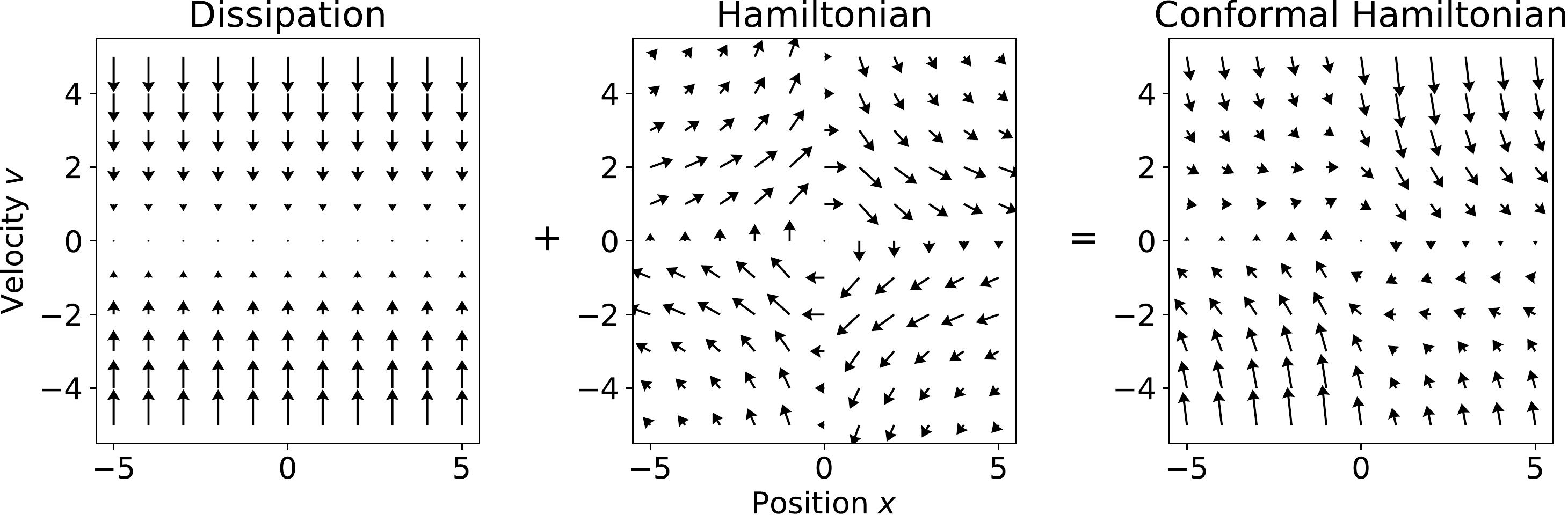} 
    \vspace{-10pt}
    \caption{Conformal Hamiltonian fields with the $\sas$ kinetic energy for $f(x) =-\log \frac{1}{\pi}\frac{1}{x^{2}+1}$. Top $\alpha=2$, bottom $\alpha=1.7$.}
    \label{fig:quiver_v2}
\end{figure}

On the other hand, we visualize the conformal Hamiltonian field generated by this dynamics in Figure~\ref{fig:quiver_v2} for $f(x) = g_{1}(x)=-\log \frac{1}{\pi}\frac{1}{x^{2}+1}$. The figure shows that conformal Hamiltonian generated by the dynamics with $\alpha=2$ has a very slow concentration behavior towards the minimum at the origin, whereas this behavior is alleviated when $\alpha=1.7$ where the field concentrates faster.

\section{Additional Experimental Results}

In this section, we provide the additional experimental results that were mentioned in the main document for width $32$, $64$, and $512$.

 \begin{figure}[t]

    \centering
    \includegraphics[width=0.7\columnwidth]{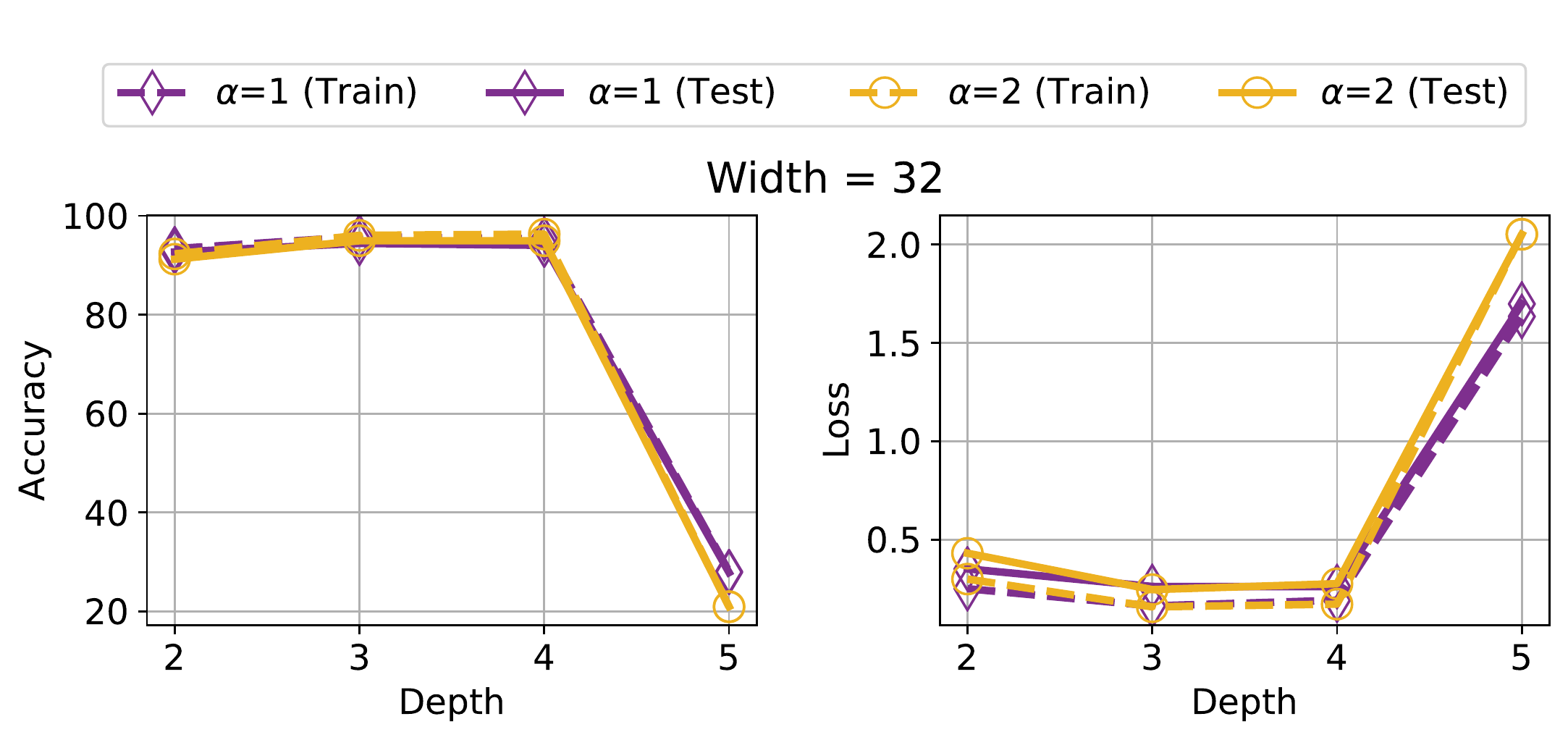} \\
    \includegraphics[width=0.7\columnwidth]{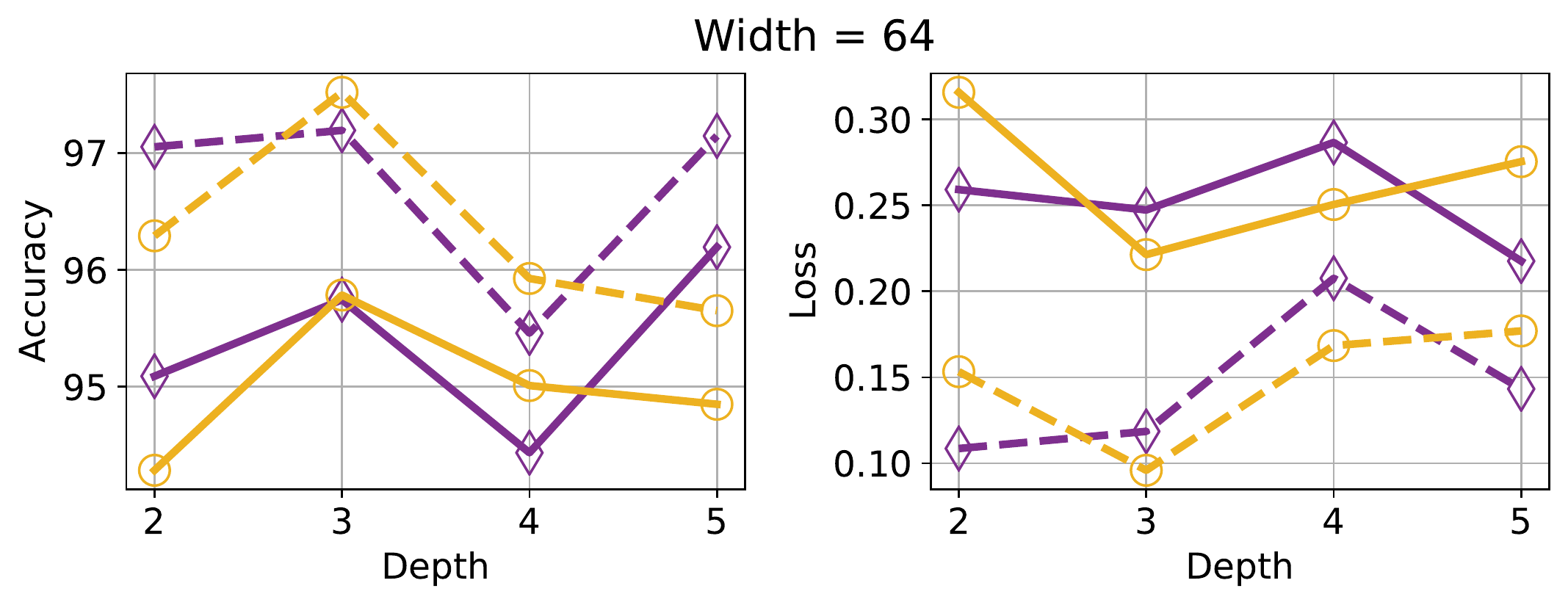} \\
    \includegraphics[width=0.7\columnwidth]{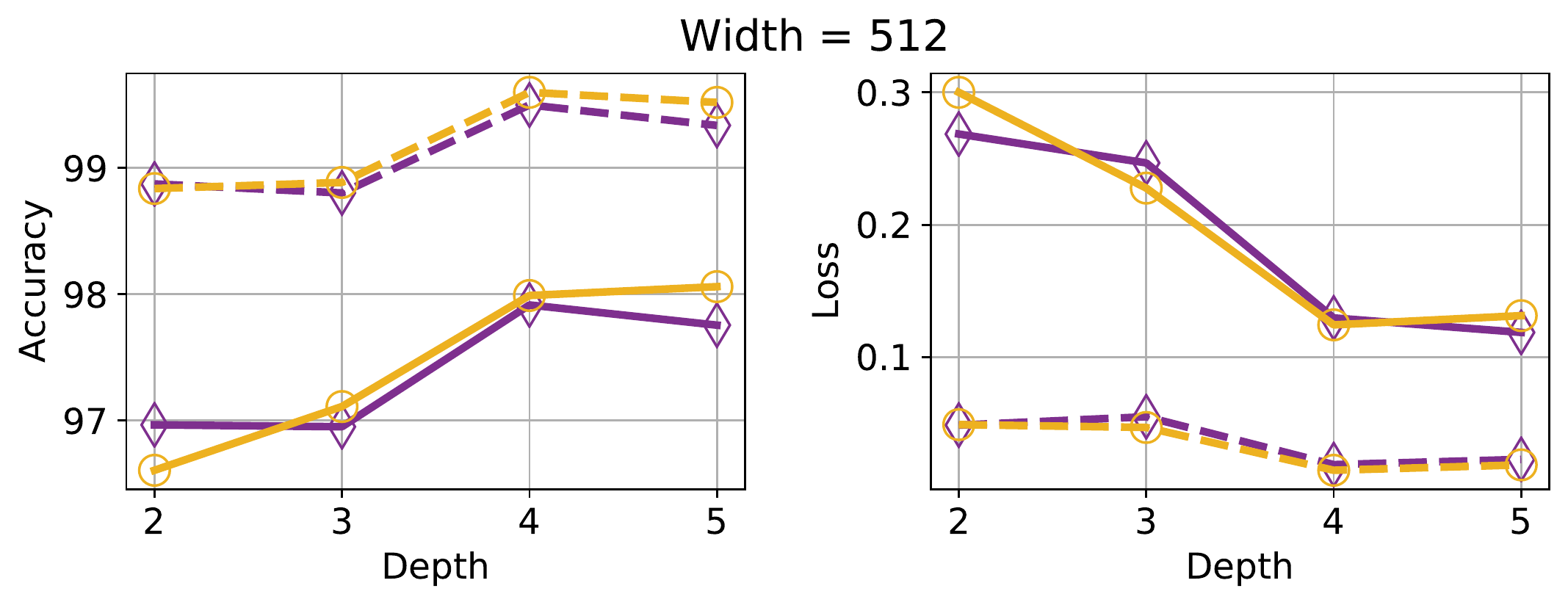}
    \vspace{-10pt}
    \caption{Neural network results on MNIST.}
    \label{fig:fcn_mn2}
\end{figure}

\end{document}